\documentclass[journal]{IEEEtran}
\usepackage{todonotes}
\usepackage{color}
\usepackage{flushend}

\ifCLASSINFOpdf
\else
\fi
\usepackage{dingbat}
\usepackage{amsmath}
\usepackage{array}
\newcolumntype{P}[1]{>{\centering\arraybackslash}p{#1}}
\usepackage{graphicx}
\ifCLASSOPTIONcompsoc
  \usepackage[caption=false,font=normalsize,labelfont=sf,textfont=sf]{subfig}
\else
  \usepackage[caption=false,font=footnotesize]{subfig}
\fi
\usepackage{url}
\usepackage{color}

\begin{document}

\title{Unsupervised and self-adaptative techniques for cross-domain person re-identification}

\author{Gabriel~Bertocco,
        Fernanda~Andal\'{o},~\IEEEmembership{Member,~IEEE,}\\
        and~Anderson~Rocha,~\IEEEmembership{Senior~Member,~IEEE}% <-this % stops a space
\thanks{Gabriel Bertocco is a Ph.D. student at the Artificial Intelligence Lab. (\textbf{Recod.ai}), Institute of Computing, University of Campinas, Brazil}% <-this % stops a space
\thanks{Fernanda~Andal\'{o} is a researcher associated to the Artificial Intelligence Lab. (\textbf{Recod.ai}), Institute of Computing, University of Campinas, Brazil}% <-this % stops a space
\thanks{Anderson Rocha is an Associate Professor and Chair of the Artificial Intelligence Lab. (\textbf{Recod.ai}) at the Institute of Computing, University of Campinas, Brazil. 
}% <-this % stops a space
\thanks{This paper has supplementary downloadable material available at http://ieeexplore.ieee.org., provided by the author. The material includes an extra file with further quantitative and qualitative analysis. Contact gabriel.bertocco@ic.unicamp.br for further questions about this work.}}



\markboth{IEEE Transactions on Information Forensics and Security,~Vol.~16, 2021}%
{Bertocco \MakeLowercase{\textit{et al.}}: Unsupervised and self-adaptative techniques for cross-domain person re-identification}



\maketitle

\begin{abstract}
Person Re-Identification (ReID) across non-overlapping cameras is a challenging task, and most works in prior art rely on supervised feature learning from a labeled dataset to match the same person in different views. However, it demands the time-consuming task of labeling the acquired data, prohibiting its fast deployment in forensic scenarios. Unsupervised Domain Adaptation (UDA) emerges as a promising alternative, as it performs feature adaptation from a model trained on a source to a target domain without identity-label annotation. However, most UDA-based methods rely upon a complex loss function with several hyper-parameters, hindering the generalization to different scenarios. Moreover, as UDA depends on the translation between domains, it is crucial to select the most reliable data from the unseen domain, avoiding error propagation caused by noisy examples on the target data --- an often overlooked problem.  In this sense, we propose a novel UDA-based ReID method that optimizes a simple loss function with only one hyper-parameter and takes advantage of triplets of samples created by a new offline strategy based on the diversity of cameras within a cluster. This new strategy adapts and regularizes the model, avoiding overfitting the target domain. We also introduce a new self-ensembling approach, which aggregates weights from different iterations to create a final model, combining knowledge from distinct moments of the adaptation. For evaluation, we consider three well-known deep learning architectures and combine them for the final decision. The proposed method does not use person re-ranking nor any identity label on the target domain and outperforms state-of-the-art techniques, with a much simpler setup, on the Market to Duke, the challenging Market1501 to MSMT17, and Duke to MSMT17 adaptation scenarios. 

\end{abstract}


% \fi
\IEEEpeerreviewmaketitle

\section{Introduction}
\IEEEPARstart{P}{erson} Re-Identification (ReID) has gained increasing attention in the last years in the Computer Vision and Forensic Science communities due to its broad range of applications for person tracking, crime investigation, and surveillance. One of the main goals when dealing with forensic problems is to answer ``who took part in an event?''. Person ReID comprises the primary techniques to find possible people, or groups of people, involved in an event and to, ultimately, propose candidate suspects for further investigation~\cite{padilha2020forensic}.

Person ReID aims to match the same person in different non-overlapping views in a camera system. Thanks to the considerable discrimination power given by deep learning, recent works~\cite{qian2017multi, sun2018beyond, zhou2019omni, chen2020salience, liu2020unity} consider supervised feature learning on a labeled dataset, which yields high values of mean Average Precision (mAP) and top Ranking accuracy.\par

However, the labeling of massive datasets demanded by deep learning is time-consuming and error-prone, especially when targeting forensic applications. In this context, Unsupervised Domain Adaptation (UDA) aims to adapt a model trained on a source dataset to a target domain without the need for identity information of the target samples. Most ReID methods that follow this approach are based on label proposing, in which feature vectors of target images are extracted and clustered. Upon unsupervised training, these clusters receive pseudo-labels for the adaptation to the target domain. 

Several works~\cite{fan2018unsupervised, song2020unsupervised, fu2019self, zhang2019self, zhai2020ad} apply the pseudo-labeling principle by developing different ways to propose and refine clusters on the target domain. The aim is to alleviate noisy labels, which can harm feature learning. Our method follows this trend, but we consider a more general clustering algorithm differently from previous work, which can relax the criteria to select data points, allowing clusters with arbitrary densities in feature space. By not forcing all clusters to have the same complexity, we can utilize the density information to better group relevant data points.

As we are dealing with data from an unknown target domain, clusters can have different degrees of reliability, i.e., contain different quantities of noisy labels. We need to select the most reliable clusters to optimize the model at each iteration of the clustering process. The generated model must also be camera-invariant to generate the same feature representation for an identity, regardless of the camera point of view. Based on these observations, we hypothesize that clusters with more cameras might be more reliable to optimize the model. Suppose that a cluster contains images of the same identity seen from two or more cameras. In this case, the model was able to embed these images close to each other in the feature space, overcoming differences in illumination, pose, and occlusion, which are inherently present in different camera vantage points. 

We argue that the greater the number of different cameras in a cluster, the more reliable this cluster is to optimize the model. Following this idea, we propose a new way to create triplets of samples in an offline manner. We select one sample as an anchor for each camera represented in a cluster and two others as positive and negative examples. As a positive example, we choose a sample from one of the other represented cameras. In contrast, the negative example is a sample from a different cluster but with the same camera as the anchor. Consequently, the greater the number of cameras in a cluster, the more diverse the triplets to train the model. With this approach, we give more importance to the more reliable clusters, regularize the model, and alleviate the dependency on hyper-parameters by using a single-term and single-hyper-parameter triplet loss function. This technique brings robustness and generability to the final model, easing its adaptation to different scenarios. \par  

Another important observation is that, at different points of the adaptation from a source to a target domain, the model holds different levels of knowledge as different portions of the target data are considered each time. Thus, we argue that the model has complementary knowledge in different iterations during training. Based on this, we propose a self-ensembling strategy to summarize the knowledge from various iterations into a unique final model.

Finally, based on recent advances in ensemble-based methods for ReID~\cite{ge2020mutual, zhai2020multiple}, we propose to combine the knowledge acquired by different architectures. Unlike prior work, we avoid complex training stages by simply assembling the results from different architectures only during evaluation time. \par 

To summarize, the contributions of our work are:

\begin{itemize}
    \item A new approach to creating diverse triplets based on the variety of cameras represented in a cluster. 
    %The greater the number of cameras in a cluster, the greater the number of cross-camera pair combinations used to train the model. 
    This approach helps the model to be camera-invariant and more robust in generating the same person's features from different perspectives. It also allows us to leverage a single-term and single-hyper-parameter triplet loss function to be optimized.
    
    \item A novel self-ensembling fusion method, which enables the final model to summarize the complementary knowledge acquired during training. This method relies upon the knowledge hold by the model in different checkpoints of the adaptation process.

    \item A novel ensemble technique to take advantage of the complementarity of different backbones trained independently. Instead of applying the typical knowledge distilling~\cite{hinton2015distilling} or co-teaching~\cite{han2018co, chen2020enhancing} methods, which add complexity to the training process, we propose using an ensemble-based prediction.   
\end{itemize}   

\section{Related Work}
\label{sec:relatedwork}
Several works address Unsupervised Domain Adaptation for Person Re-Identification. They can be roughly divided into three categories: generative, attribute alignment, and label proposing methods.

\subsection{Generative Methods}
ReID generative methods aim to synthesize data by translating images from a source to a target domain. Once data from the source dataset is labeled, the translated images on the target context receive the same labels as the corresponding original images. The main idea is to transfer low- and mid-level characteristics from the target domain, such as background, illumination, resolution, and even clothing, to the images in the source domain. These methods create a synthetic dataset of labeled images with the same conditions as the target domain. And to adapt the model, they apply supervised training. Some works in this category are SPGAN~\cite{deng2018image}, PTGAN~\cite{wei2018person}, AT-Net~\cite{liu2019adaptive}, CR-GAN~\cite{chen2019instance}, PDA-Net~\cite{li2019cross}, and HHL~\cite{zhong2018generalizing}. Besides transferring the characteristics from source to target domain for image-level generation, DG-Net++~\cite{zou2020joint} also applies label proposing through clustering. The final loss is the aggregation of the GAN-based loss function to generate images, along with the classification loss defined for the proposed labels. By doing this, they perform the disentangling and adaptation of the features on the target domain. 

%\textcolor{red}{%(R3C4)
CCSE~\cite{lin2020unsupervisedccse}
performs camera mining and, using a GAN-based model, generates synthetic data for an identity considering the point of view of each other camera, increasing the number of images available for training. They leverage new clustering criteria to avoid creating massive clusters comprising most of the dataset and potentially having two or more true identities assigned to the same pseudo-label. 
%They continuously merge clusters to group images of the same identity during the training.
Finally, they train directly from ImageNet, without considering any specific source domain.
%Finally they do not use any person re-identification source domain and leverage the adaptation with a backbone trained directly from ImageNet. 
In comparison, our solution does not require synthetic images since we explore the cross-camera information inside each cluster using only real images. This leads our method to outperform CCSE considering the same training conditions (unsupervised scenario). %}

%More specifically, if the target dataset has $C_{t}$ cameras and an image of an identity has been captured by a camera $i \in \{1,...,C_{t}\}$, they generate a synthetic image for each camera $j \in \{1,..,C_{t}\}$ with $i \neq j$ for this identity. Then, each image will have more $C_{t} - 1$ copies and the dataset is increased by $C_{t}$. Moreover, they leverage a new clustering criteria to avoid creating huge clusters comprising the majority of the dataset and potentially having two or more true identities assigned to the same pseudo-label. They continuously merge clusters to group images of the same identity during the training. Finally they do not use any person re-identification source domain, and leverage the adaptation with a backbone trained directly from ImageNet. In comparison, our solution does not require creating synthetic images among cameras since we directly explore the cross-camera information inside each cluster using only real/actual images. This avoids adding noise and complexity to the model training and allows a better knowledge over the target domain. Our model outperforms CCSE considering the same training conditions (Unsupervised Scenario), where we do not use pre-training in any person reid-related source domain nor images generation as will show on Section \ref{sec:experiments}}.

\subsection{Attribute Alignment Methods} 
These methods seek to align common attributes in both domains to ease transferring knowledge from source to target. Such features can be clothing items (backpacks, hats, shoes) and other soft-biometric attributes that might be common to both domains. These works align mid-level features and enable the learning of higher semantic features on the target domain. Works such as TJ-AIDL~\cite{wang2018transferable} consider a fixed set of attributes. %(23 to 27 attributes).
However, source and target domains can have substantial context differences, leading to potentially different attributes. For example, the source domain could be recorded in an airport and the target domain in a shopping center. To obtain a better generalization, in~\cite{lin2018multi}, the authors propose the Multi-task Mid-level Feature Alignment (MMFA) technique to enable the method to learn attributes from both domains and align them for a better generalization on the target domain. Other methods, such as  UCDA~\cite{qi2019novel} and CASCL~\cite{wu2019unsupervised}, aim to align attributes by considering images from different cameras on the target dataset. 

\subsection{Label Proposing Methods} 
Methods in this category predict possible labels for the unlabeled target domain by leveraging clustering methods (K-means \cite{lloyd1982kmeans}, %Agglomerative Clustering~\cite{Zepeda-Mendoza2013}, 
DBSCAN~\cite{ester1996density}, among others). Once the target data is pseudo-labeled, the next step is to train models to learn discriminative features on the new domain. PUL~\cite{fan2018unsupervised} applies the Curriculum Learning technique to adapt a model learned on a source domain to a target domain. However, as K-means is used to cluster the features, it is not possible to account for camera variability. As K-means generates only convex clusters, it cannot find more complex cluster structures, hindering the performance. UDAP~\cite{song2020unsupervised} and ISSDA-ReID~\cite{tang2019unsupervised} utilize DBSCAN as the clustering algorithm along with labeling refinement.
%\textcolor{red}{%(R3C4) - 
SSG~\cite{fu2019self} also applies DBSCAN to cluster features of the whole, upper, and low-body parts of identities of interest. The final loss is the sum of individual triplet losses in each feature space (body part). Similar to our work, they use a source domain to pre-train the model and the target domain for adaptation. However, they do not perform cross-camera mining, cluster filtering, nor ensembling. These elements of our solution allow it to outperform SSG in all adaptation scenarios. %}. 
\par  

ECN~\cite{zhong2019invariance}, ECN-GPP~\cite{zhong2020learning}, MMCL~\cite{wang2020unsupervised}, and Dual-Refinement~\cite{dai2020dual} use a memory bank to store features, which is updated along the training to avoid the direct use of features generated by the model in further iterations. The authors aim to avoid propagating noisy labels to future training steps, contributing to keeping and increasing the discrimination of features during training. \par  

PAST~\cite{zhang2019self} applies HDBSCAN~\cite{campello2013density} as the clustering method, which is similar to OPTICS~\cite{ankerst1999optics} --- the algorithm of choice in our work. However, the memory complexity of OPTICS is $O(n)$, while for HDBSCAN is $O(n^2)$, making our model more memory efficient in the clustering stage.   \par 

MMT~\cite{ge2020mutual}, MEB-Net~\cite{zhai2020multiple}, ACT~\cite{yang2020asymmetric}, SSKD~\cite{yin2020sskd}, and ABMT~\cite{chen2020enhancing} are ensemble-based methods. They consider two or more networks and leverage mutual teaching by sharing one network's outputs with the others, making the whole system more discriminative on the target domain. However, training models in a mutual-teaching regime brings complexity in memory and to the general training process. Besides that, noisy labels can be propagated to other ensemble models, hindering the training process. Nonetheless, ensemble-based learning provides the best performance among state-of-art methods. We propose using ensembles only during inference to simultaneously eliminate the complexity added to the training, still taking advantage of knowledge complementary between the models. 

Our work is also based on Curriculum Learning with Diversity~\cite{jiang2014self}, a schema whereby the model starts learning with easier examples, i.e., samples that are correctly classified with a high score early in training. However, in a multi-class problem, one of the classes might have more examples correctly classified early on, making it easier than the other classes. Therefore, in Curriculum Learning with Diversity, the method selects the most confident samples (easier samples) from the easier classes, including some examples from the harder ones. In this way, it enables the model to learn in an easy-to-hard manner, avoiding local minima and allowing better generalization.

Even though recent work achieves competitive performances, there are some limitations that we aim to address in our work. First, generative methods bring complexity by considering GANs to translate images from a domain to the other. Second, attribute Alignment methods only tackle the alignment of low and mid-level features. Third, methods in both categories need images from source and target domains during adaptation. Finally, the last Label Proposing methods consider mutual-learning or co-teaching, which brings complexity to the training stage.

%\textcolor{red}{%(R3C2)
Similarly, we assume to have only camera-related information, i.e., we know from which camera (viewpoint) an image was taken. In all steps, we use pseudo-identity information exclusively given by the clustering algorithm without relying on any ground-truth information. We differ from the prior art by using a new diversity learning scheme and generating triplets based on each cluster's diversity of points of view. As we train the whole model, the method also learns high-level features on the target domain. We simplify the training process by considering one backbone at a time, without mutual information exchange during adaptation. Finally, we apply model ensembling for inference after the training process.%}

\section{Proposed Method}
\label{sec:proposed_method}

\begin{figure*}[ht]
\centering
\includegraphics[width=6.5in]{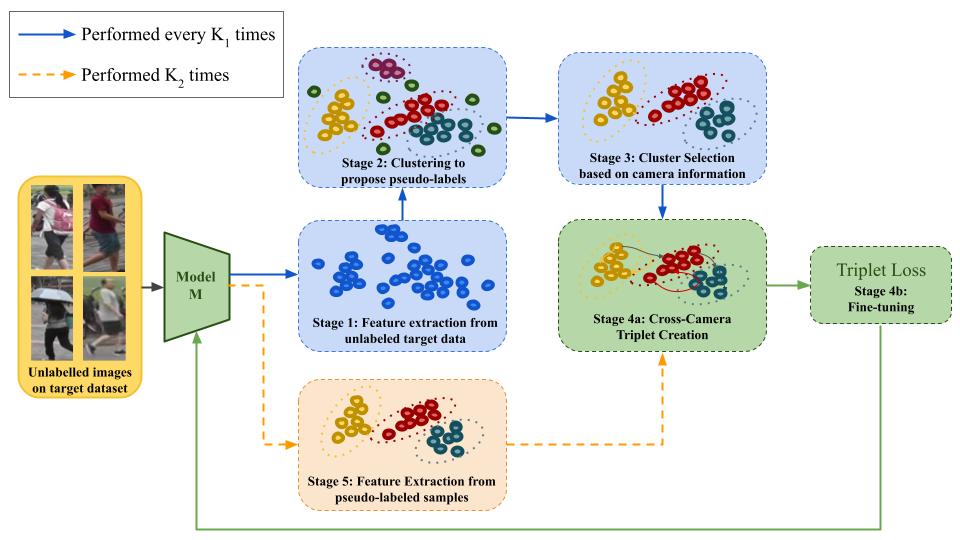}
\caption{Overview of the training phase. %\textcolor{red}{%(R3C2) - 
We assume to have camera-related information, i.e., we know the camera used to acquire each image; and we do not rely on any ground-truth label information about the identities on the target domain. %} 
The pipeline has two flows: the blue flow is executed every $K_{1}$ times, and the orange flow is executed $K_{2}$ times. Both flows share steps in green. In Stage 1, we initially extract feature vectors for each training image in the target domain using model $M$, %\textcolor{red}{%(R3C2) - 
and cluster them using the OPTICS algorithm in Stage 2 to propose pseudo-labels. %} 
Afterward, we perform cluster selection in Stage 3, removing outliers and clusters with only one camera. Then, triplets are created based on each cluster's diversity in Stage 4a and used to train the model in Stage 4b. These steps are denoted by the blue flow in which the Clustering and Cluster Selection are performed. Instead of going back to Stage 1, the method follows the orange flow. In Stage 5, we extract feature vectors of the samples selected in Stage 3, and the process continues to Stage 4a and 4b again. The blue flow marks an iteration, while the orange flow is called an epoch. Therefore, in each iteration, we have $K_{2}$ epochs.}

\label{fig:overview_pipeline}
\end{figure*}

Our approach to Person ReID comprises two phases: training and inference.  Figure~\ref{fig:overview_pipeline} depicts the 
training process, while 
%\textcolor{red}{
Table~\ref{tab:terminology} shows %(R2C1)
the variables used in this work.%} 

\begin{table}[ht]
\caption{%\textcolor{red}{
Variables' meaning in this work} %}
\label{tab:terminology}
\centering
\begin{tabular}{|p{1.0cm}|p{6.0cm}|}
\hline
\hline
Variable & Meaning  \\ \hline
$n_{b}$ & Number of different backbones in the Ensemble \\
$M$ & Model backbone \\ 
$K_{1}$ & Number of iterations of the blue flow in Figure \ref{fig:overview_pipeline} \\
$K_{2}$ & Number of iterations of the orange flow in Figure \ref{fig:overview_pipeline} \\
$c_{i}$ & i-th cluster in the feature space \\
$n_{i}$ & Number of cameras in cluster $c_{i}$ \\
$cam_{j}$ & j-th camera in a cluster \\
$x_{i}^{s}$ & i-th image in the source domain \\
$x_{i}^{t}$ & i-th image in the target domain \\
$y_{i}^{s}$ & Label of the i-th image in the source domain \\
$N_{s}$ & Number of images in the source domain \\
$N_{t}$ & Number of images in the target domain \\
$m$ & Number of anchors per camera in a cluster \\
$\alpha $ & Margin parameter of the Triplet Loss \\
$B$ & Batch of triplets in an iteration \\

\hline
\end{tabular}
\end{table}

During training, we independently optimize $n_{b}$ different backbones to adapt the model to the target domain. This phase is divided into five main stages that are performed iteratively: feature extraction from all data; clustering; cluster selection; cross-camera triplet creation and fine-tuning; feature extraction from pseudo-labeled data.

After training, we perform the proposed self-ensembling phase to summarize the training parameters in a single final model based on the weighted average of model parameters from each different checkpoint. We perform this step for each backbone independently and, in the end, we have $n_{b}$ self-ensembled models. 

During inference, for a pair query/gallery image, we calculate the distance between them considering feature vectors extracted by each of the $n_{b}$ models. Hence, for each query/gallery pair, we have $n_{b}$ distances, one for each of the trained models. We then apply our last ensemble technique: the $n_{b}$ distances are averaged to obtain a final distance. Finally, based on this final distance, we take the label of the closest gallery image as the query label.

\subsection{Training Stages 1 and 2: Feature Extraction from all data and Clustering}
Let $ D^{s} = \{(x_{i}^{s}, y_{i}^{s})\}_{i=1}^{N_{s}} $ be a labeled dataset representing the source domain, formed by $ N_{s} $ images $x_{i}^{s}$ and their respective identity labels $y_{i}^{s}$; and  
let $ D^{t} = \{(x_{i}^{t})\}_{i=1}^{N_{t}} $ be an unlabeled target dataset representing the target domain, formed by $ N_{t} $ images $x_{i}^{t}$.  Before applying the proposed pipeline, we firstly train a model $M$ in a supervised way, with source dataset $D^{s}$ and its labels. After training, assuming source dataset $ D^{s}$ is not available anymore, we perform transfer learning, updating $M$ to the target domain, only considering samples from unlabeled target dataset $ D^{t} $.  

With model $M$ trained on $D^{s}$, we first extract all feature vectors from images in $ D^{t} $ and create a new set of feature vectors $\{ M(x_{i}^{t})\}_{i=1}^{N_{t}}$.
We remove possible duplicates by checking if there is a replacement from one of them, which might be caused by duplicate images on target data. The remaining feature vectors are L2-normalized to embed them into a unit hypersphere. The normalized feature vectors are clustered using the OPTICS algorithm to obtain pseudo labels. 

The OPTICS algorithm~\cite{ankerst1999optics} leverages the principle of dense neighborhood, similarly to DBSCAN~\cite{ester1996density}. DBSCAN defines the neighborhood of a sample as being formed by its closest feature vectors, with distances lower than a predefined threshold. Clusters are created based on these neighborhoods, and samples not assigned to any cluster are considered outliers. If the threshold changes, other clusters are discovered, current clusters can be split or combined to create new ones.  In other words, if we change the threshold, other clusters might appear, creating a different label proposing for the samples. However, clusters that emerge from real labels often have different distributions and densities, indicating that a generally fixed threshold might not be sufficient to detect them. In this sense, OPTICS relaxes DBSCAN by ordering feature vectors in a manifold based on the distances between them, which allows the construction of a \textit{reachability plot}. Probable clusters with different densities are revealed as valleys in this plot and can be detected by their steepness. With this formulation, we are more likely to propose labels closer to real label distribution on the target data.

%the true clusters (i.e. the cluster formed if the labels were known) might have different densities
%in the feature space, so it would be interesting to select a specific threshold value for each of the identities on the target domain. In this way, we would be closer to find the true identities following their particular densities on the feature space. However, the original DBSCAN formulation allow us to define only a fixed threshold to find all clusters. In this sense, OPTICS relax this idea by ordering the points on the feature space based on the distance between them. After that they obtained the called \textit{reachability plot} and based on the steepness of different parts present on it, they extract the clusters. The steepness is a hint to detect the cluster in different densities in the feature space, which allow us to be closer to real label distribution of the target data. We refer the reader to see \cite{ankerst1999optics} for more details. 

\subsection{Training Stage 3: Cluster Selection}
\label{subsec:cluster_selection}

After the first and second stages, feature vectors are either assigned to a cluster or considered outliers. As people can be captured by one or more cameras in a ReID system, the produced clusters are naturally formed by samples acquired by different devices. We hypothesize that clusters with samples obtained by two or more cameras are more reliable than clusters with only one camera. 

If an identity is well described by model $M$, its feature vectors should be closer in the feature space regardless of the camera. Therefore, clusters with only one camera might be created due to bias to a particular device or viewpoint, and different identities captured by the same camera can be assigned to the same cluster. Besides, if a feature vector is predicted as an outlier by the clustering algorithm, it means that it does not have a good description of its image identity to be assigned to a cluster.

Based on these observations and for optimization purposes, we filter the feature vectors by discarding outliers and clusters with a single camera type.
%\textcolor{red}{%(R2C2) - 
With camera-related information, it is possible to count the number of images from each camera in a cluster. If all samples in a cluster come from the same camera, it is removed from the feature space. By doing this, we keep in the feature space only clusters with images from at least two cameras.%}
%It is important to note we assume that we have only the camera-related information available for each bounding box image in the target domain, without knowing any identity-related label. Then we are able to count the number of images from each camera to obtain how many samples from each one the cluster contains. If all samples on the cluster come from the same camera, then we say that this cluster has only one camera and it is removed from the feature space. By doing this, we keep in the feature space only clusters with images from at least two cameras. These assumptions are also present in some existing work in the prior art, such as in AD-Cluster, ECN-GPP and CCSE}.
Figure~\ref{fig:overview_pipeline} depicts this process, from Stage 2 to Stage 3, in which the outlier samples (green points) and clusters with only one camera (magenta points) are removed from the feature space. 

The remaining clusters (the ones with two or more cameras) are considered reliable to fine-tune model $M$. Furthermore, different clusters have different degrees of reliability based on the number of represented cameras. 
Suppose images captured by several cameras form a cluster. In that case, it means model $M$ can embed samples of the corresponding identity captured by all of these cameras in the feature space, eliminating point-of-view bias. In contrast, the fewer images from different points of view, the more complex the identity definition.
In this sense, we propose a new approach of creating cross-camera triplets of samples to optimize the model by emphasizing cluster diversity and forcing samples of the same identity to be closer in the feature space regardless of their acquisition camera.

%\textcolor{red}{(R3C2)It is important to remember that in all steps, we used pseudo-identity information exclusively given by the clustering algorithm without relying on any ground-truth information.}

% Clusters with more cameras mean that more features vectors from different points of view are localized closer in the feature-space, in other words, the model $ M(\cdot|\theta)$ is able to well describe the images from the same identity independent of the point of views, and put their feature vectors closer to form a cluster. Therefore those images have a high probability to be from the same identity. In counterpart, the less the amount of images from different points of view, the harder is the identity to be described. In this sense, we propose a new way to create the triplets of images to optimize the model in order to give more attention to the clusters with more diversity.

\subsection{Training Stage 4: Cross-Camera Triplet Creation and Fine-tuning}

\begin{figure}[ht]
\centering
\includegraphics[width=4.1in]{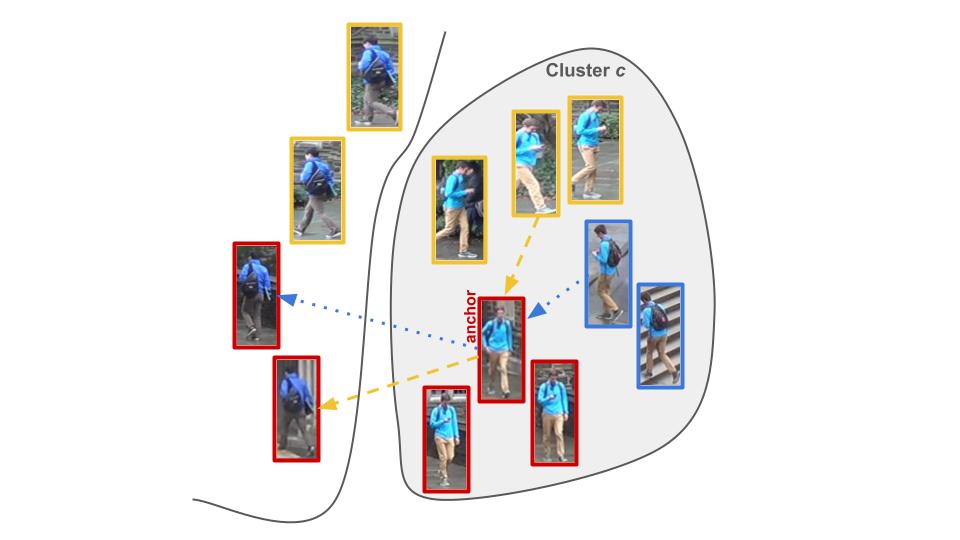}
\caption{Cross-Camera Triplet Creation. For each selected cluster, we have at least two cameras. Suppose the represented cluster $c$ has images from three cameras (represented with red, blue, and yellow contours). For each camera, we select $m$ anchors. For each anchor, we create triplets with a positive sample from other cameras in the same cluster and a negative sample with the same camera in other clusters. For instance, for camera red, we select an anchor and we sort, based on the distance, all feature vectors from cameras yellow and blue. Then we select the median feature vector from each one (represented by the arrows coming to the anchor). To select the negative sample, we sort all feature vectors from the same camera but from a cluster $\neq c$, and we choose the closest and not previously selected sample. For the triplet with a yellow median sample as positive, we select as negative the closest sample to the red anchor from another cluster (represented by the yellow arrow leaving the anchor). For the triplet with a blue median sample as positive, we select the second closest feature vector to the red anchor from another cluster (since the first closest has already been picked). This explanation assumes $m = 1$ and is repeated for cameras yellow and blue.}
\label{fig:cross_camera_triplet_creation}
\end{figure}

Figure~\ref{fig:cross_camera_triplet_creation} shows the triplet creation process. A triplet is formed by an anchor, a positive, and a negative sample. 
During optimization, the distance from the anchor to the positive sample should be minimized, while the distance to the negative sample should be maximized. Ideally, positive and negative samples should be hard-to-classify examples for the current model $M$ as easy examples do not bring diversity to the learning process. 

We initially select, as the anchor, one random sample in cluster $c$ captured by camera ${cam}_j$. For each camera ${cam}_k \neq {cam}_j$ in cluster $c$, we sort all feature vectors from camera ${cam}_k$ based on their distance to the anchor. The positive sample is then selected as being the median feature vector. The median is considered instead of the farthest sample (the hardest example) to avoid selecting a noisy example. We do not choose an easy example (the closest one) to avoid slowing down the model convergence or even getting stuck on a local minimum.
To select the negative sample, we first sort all feature vectors from camera ${cam}_j$ belonging to other clusters $\neq c$ based on their distance to the anchor. As the negative sample, we pick the closest feature vector that has not been assigned yet to a triplet. In this way, we avoid selecting the same negative sample, which brings diversity to the triplets and alleviates the harmful impact if one of the negative samples shares the anchor's same real identity.

% Let it is assume a cluster $CL_{i}$ with three cameras as shown on Figure \ref{fig:cross_camera_triplet_creation}. To create a triplet, we firstly select one of the three available cameras and randomly pick one of its images as an anchor. For each of the other cameras, we sorted the feature vectors of the images based on the distance to the selected anchor, and we pick the median feature vector as a positive sample. In this way, we avoid to select a noisy example on the cluster by not picking the furthest example that could hinder the training performance. As long as we do not choose an easy example by choosing the closest one, which could slow down the model convergence or even get stuck on a local minima early on the training phase. To select the negative sample on the triplet, we sorted all other images from different clusters but with the same camera ID of the anchor, then we select the closest one that has not been assigned to any other triplet yet. In other words, if the closest negative sample has already been assigned to a triplet, we choose the second closest, if it also has  been assigned to a triplet, then we select the third closest, and so on. In this way, we avoid to always select the same negative sample which brings diversity to the triplets, and in the same time alleviate the bad impact if one the negative samples belong to the same identity of the anchor. 

For a cluster $c_i$ with $n_i$ cameras, we generate a total of $n_i - 1$ triplets with the same anchor. If we select $m$ anchors for one camera in $c_i$, a total of $ m (n_i - 1)$ triplets are created. 
Considering that this process is repeated for each camera in $c_i$, we have a total of $n_i m (n_i - 1)$ triplets for cluster $c_i$.
Note that the triplets are created in an offline manner. The offline creation enables us to choose triplets considering a global view of the target data instead of creating them in a batch, which would bring a limited view of the target feature space. 

%However, its implementation for batches of different sizes is straightforward. 

% Based on this explanation, if a cluster $CL_{i}$ has $ C_{i} $ cameras, we generate a total of $ C_{i} - 1 $ triplets with the same anchor selected from one of the cameras. If we choose $ N $ anchor for each camera on the cluster, we have a total of $ N*(C_{i}-1)$ triplets created. Considering that the whole process is repeated for each camera on cluster, we have a total of $C_{i}*N*(C_{i}-1)$ triplets generated for the cluster. It is important to point out that the triplets are created in an offline manner. This enable us to choose triplets considering a global view of the target data, instead of choosing them in a batch that brings only a local view of the target feature space. 

The number $m$ of anchors of a camera is the same for all clusters. Consequently, the number of triplets generated for a cluster $c_i$ is $\mathcal{O}(n_i^2)$. The greater the diversity of cameras in a cluster, the greater its representativeness on the triplets. By emphasizing the clusters with more camera diversity during training, the model learns from easy-to-hard identities and is more robust to different viewpoints. In our experiments we set $m = 2$ for all adaptation scenarios.

Due to this new approach of creating cross-camera triplets, we can optimize the model by using the triplet loss~\cite{schroff2015facenet} without the need for weight decay or any other regularization term and hyper-parameters. This also suggests that cross-camera triplets help to regularize the model during training.

% In our work, we define $ N $ as the same for all cameras in all clusters, then the number of triplets generated by a cluster $ CL_{i}$ is $\mathcal{O}((CL_{i})^2)$. Therefore, the greater is the number of cameras in a cluster, the greater is the number of triplets generated by that cluster. In other words, the greater is the diversity of the cluster, the greater will be its representativity on the triplets to fine-tune the model. This enforces the model to pay more attention to the cluster with more diversity, which allows the model to start learning from easier identities, as explained on the beginning of this section, along with be more robust for different point of views. This brings a strong robustness to our model that allows us to optimize it using only triplet loss without adding any other regularization term or needing to define a hyperparamter on loss function. Even weight decay is set to zero and  our model is able to train and get more and more discriminant on target domain as will show later on experiments section. This also demonstrates that our new way to create triplets also brings regularization by it own during model training. This new process to create triplets we call by Cross-Camera Triplet Creation.     

After creating the triplets in an offline manner, we optimize the model using the standard triplet loss function:

\begin{equation}
\label{eq:LWHTL}
    L = \frac{1}{|B|}\sum_{(x_{a}, x_{p}, x_{n}) \in B} \left[d(x_{a}, x_{p}) - d(x_{a}, x_{n}) + \alpha \right]_{+},
\end{equation}

\noindent where $ B $ is a batch of triplets, $x_{a}$ is the anchor, $x_{p}$ is the positive sample and $x_{n} $ is the negative one. $ \alpha $ is the margin that is set to $0.3$ and $[.]_{+}$ is the $max(0, .)$ function. This is illustrated in Figure \ref{fig:overview_pipeline}, Stage 4b.  

\subsection{Stage 5: Feature Extraction from Pseudo-Labeled Samples}

This stage is part of the orange flow performed after Fine-tuning (Stage 4b). The main idea is to keep the pseudo-labeled clusters from Stage 3, recreating a new set of triplets based on the new distances between samples after the model update in Stage 4b, bringing more diversity to the training phase. To do so, we extract feature vectors only for samples of the pseudo-labeled clusters selected in Stage 3. The orange flow is performed $K_{2}$ times, and a complete cycle defines an epoch. The blue flow is performed every $K_{1}$ times, and a complete cycle defines an iteration. Therefore, in each iteration, we have $K_{2}$ epochs. This concludes the training phase. 

Unlike the five best state-of-the-art methods proposed in the prior art (DG-Net++, MEB-Net, Dual-Refinement, SSKD, and ABMT), our solution is trained with a single-term loss, which contains only one hyper-parameter. Even the weight decay has been removed, as the proposed method can already calibrate the gradient to avoid overfitting, as we show in Section~\ref{sec:experiments}. Moreover, prior work performs clustering on the training phase through k-reciprocal Encoding~\cite{zhong2017re}, which is a more robust distance metric than Euclidean distance. However, it has a higher computational footprint, as it is necessary to check the neighborhood of each sample whenever distances are calculated. For training simplicity, we opt for standard Euclidean distance to cluster the feature vectors. However, as k-reciprocal encoding gives the model higher discrimination, we adopt it during inference time. Therefore, different from previous works, we calculate k-reciprocal encoding only once during inference.

\subsection{Self-ensembling}
\label{sec:network_fusion}
\begin{figure}[ht]
\centering
\includegraphics[width=3.0in]{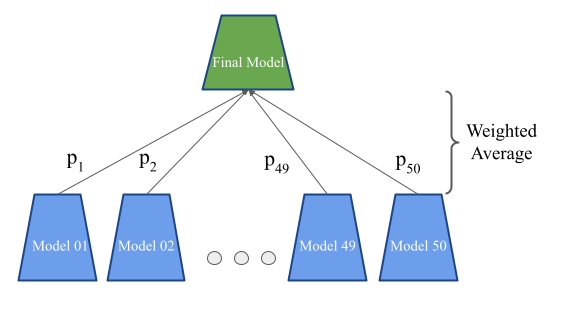}
\caption{Self-Ensembling scheme after training. Different amounts of the target data (with no label information whatsoever) are used to fine-tune the model during the adaptation process. Different models created along the adaptation can be} complementary. We create a new final model by weight averaging the models' parameters from different iterations. Weight $p_{i}$ is based on the amount of reliable data from the target domain on the i-th iteration. We end up with a single model encoding knowledge from different moments of the adaptation.
\label{fig:network_surgery}
\end{figure}

 Our last contribution relies upon the curriculum learning theory. Different iterations of the training phase consider different amounts of reliable data from the target domain, as shown in Section~\ref{sec:experiments}. This property leads us to hypothesize that knowledge obtained at different iterations is complementary. Therefore, we propose to summarize knowledge from different moments of the optimization in a unique final model. However, as the model discrimination ability increases as more iterations are performed (the model is able to learn from more data), we propose combining the model weights of different iterations by weighting their importance with the amount of reliable data used in the corresponding iteration. We perform this weighted average of the model parameters as:

\begin{equation}
\label{eq:weight_average}
    \theta_{final} = \frac{\sum\limits_{p \in P} p_{i}.\theta_{i}}{\sum\limits_{p \in P} p_{i}},
\end{equation}

\noindent where $ \theta_{i} $ represents the model parameters after the i-th iteration and $p_{i} $ is the weight assigned to $ \theta_{i} $. Weight $ p_{i} $ is obtained based on the reliability of the target domain; if more data from the target domain is considered in an iteration, it means that the model is more confident, and then it can have more discrimination power on the target domain. Hence, $p_{i}$ is equal to the percentage of reliable target data in the i-th iteration. Consequently, a model that takes more data from the target to train will have a higher weight $ p_{i} $. Self-Ensembling is illustrated in Figure~\ref{fig:network_surgery}. Note that we directly deal with the model's learned parameters and create a new one by averaging the weights.

We end up with a single model containing a combination of knowledge from different adaptation moments, which significantly boosts performance, as shown in Section~\ref{sec:ablation_study}.

\subsection{Ensemble-based prediction}
After training and performing the self-ensemble fusion, we have a single model adapted from the source to the target domain. However, due to the high performance of ensemble-based methods in recent ReID literature \cite{ge2020mutual, zhai2020multiple}, as a last measure, we leverage a combination of $n_{b}$ different architectures to make a final prediction considering even more learned knowledge, which improves performance on the target dataset. We apply the ensemble technique only for inference, different from~\cite{ge2020mutual, zhai2020multiple} that leverage a mutual-teaching regime on training time. In turn, we avoid bringing complexity to the training but still take advantage of the complementarity from different architectures during inference. 

To perform the ensemble-based prediction, we first calculate the feature distance of the query to each image on gallery for each of the $n_{b}$ final models. Let $ f_{k}(x) = M_{k}(x)$ be the L2-normalized feature vector of image $ x $ obtained with model $ M_{k}$ and $d(f_{k}(q), f_{k}(g_{i}))$ be the distance between the feature vectors of the query $ q $ and of the i-th image gallery $ g_{i} $ extracted using the k-th model on ensemble. The final distance between query $ q $ and gallery image $ g_{i} $ is given by:

\begin{equation}
\label{eq:ensemble_backbones}
    d_{final}(q, g_{i}) = \frac{1}{K} \sum_{k = 1}^{K} d(f_{k}(q), f_{k}(g_{i})),
\end{equation}
\noindent where $ K $ is the number of models in the ensemble. In this way, we can incorporate knowledge from different models encoded as the distance between two feature vectors. After obtaining the distance between query $ q $ and all images in the gallery, we take the label of the closest gallery image as the query label. 

%\textcolor{red}{%(R2C3) -
We consider an equal contribution from each backbone. Without labels on the target domain, it is impossible to evaluate the impact of the individual models and give them proportional weights on the combination. %} %For instance, which of them obtained the best performance and then assign it the highest weight.} 

%\textcolor{red}{We could measure the compactness of the generated clusters for each model backbone after self-ensembling and give a stronger weight to the one with lowest cluster compactness, as seen on \cite{zhai2020multiple} with the proposed Authority Regularization. However, in our case, different backbones evolve in different ways during the training process, which leads to different clusters, indicating different knowledge acquired along the training. This can be seen in Figures \ref{fig:reliability_market2duke} and \ref{fig:reliability_duke2market} where we show the reliability progress for each backbone on the target dataset. We see that OSNet ends with higher reliability than ResNet50 and DenseNet121, which means different amounts of samples and potentially different generated clusters in the last iteration. In this case, the cluster compactness would be calculated considering different samples and clusters, which could lead to an unfair comparison among backbones. Therefore we do not adopt any kind of further processing to combine the backbones and prefer a simple average to obtain the final distance between query and a gallery image.}   

\section{Experiments and Results}
\label{sec:experiments}

This section presents the datasets we adopt in this work and compares the proposed method with the prior art with a comprehensive set of experiments considering different, and challenging, source/target domains.

\subsection{Datasets}
\label{subsec:datasets}
To validate our pipeline, we used three large-scale benchmark datasets present on Re-ID literature:

\begin{itemize}
    \item Market1501 \cite{zheng2015scalable}: It has 12,936 images of 751 identities in the training set and 19,732 images in the testing set. The testing set is still divided into 3,368 images for the query set and 15,913 images for the gallery set. Following previous work, we removed the ``junk'' images from the gallery set, so 451 images are discarded. This dataset has a total of six non-overlapping cameras. Each identity is captured by at least two cameras.
    
    \item DukeMTMC-ReID \cite{ristani2016performance}: It has 16,522 images of 702 identities in the training set and 19,889 images in the testing set. The testing set is also divided into 2,228 query images and 17,661 gallery images of other 702 identities. The dataset has a total of eight cameras. Each identity is captured by at least two cameras.

    \item MSMT17 \cite{wei2018person}: It is the most challenging ReID dataset present in the prior art. It comprises 32,621 images of 1,401 identities in the training set and 93,820 images of 3,060 identities in the testing set. The testing set is divided into 11,659 images for a query set and 82,161 images for a gallery set. It comprises 15 cameras recorded in three day periods (morning, afternoon, and noon) on four different days. Besides, of the 15 cameras, 12 are outdoor cameras, and three are indoor cameras. Each identity is captured by at least two cameras. 
    
\end{itemize}

As done in previous work in the literature, we remove from the gallery images with the same identity and camera of the query to assess the model performance in a cross-camera matching. Feature vectors are L2-normalized before calculating distances. For evaluation, we calculate the Cumulative Matching Curve (CMC), from which we report Rank-1 (R1), Rank-5 (R5), and Rank-10 (R10), and mean Average Precision (mAP).

\subsection{Implementation details}
\label{subsec:implementation_details}
In terms of deep-learning architectures, we adopt ResNet50~\cite{he2016deep}, OSNet~\cite{zhou2019omni}, and DenseNet121~\cite{huang2017densely}, i.e., $n_{b} = 3$, all of them pre-trained on ImageNet~\cite{deng2009imagenet}. To test them on an adaptation scenario, we choose one of the datasets as the source and another as the target domain. We train the backbone over the source domain and the adaptation pipeline over the target domain. We consider Market1501 and DukeMTMC-ReID as source domains, leaving MSMT17 only as the target dataset (the hardest one in the prior art). This way, we have four possible adaptation scenarios: Market $\rightarrow$ Duke, Duke $\rightarrow$ Market, Market $\rightarrow$ MSMT17, and Duke $\rightarrow$ MSMT17. We keep those scenarios (without MSMT17 as a source) to have a fair comparison with state-of-the-art methods. Besides, the most challenging scenario is MSMT17 as the target dataset: we train backbones on simpler datasets (Market and Duke) and adapt their knowledge to a harder dataset, with almost the double number of cameras and with many more identities recorded in different moments of the day and the year. This enables us to test the generalization of our method in adaptation scenarios where source and target domain have substantial differences in the number of identities, camera recording conditions, and environment. 

We used the code available at~\cite{torchreid} to train OSNet and at~\cite{zhai2020multiple} to train ResNet50 and DenseNet121 over the source domains. Our source code is based on PyTorch~\cite{NEURIPS2019_9015} and it is freely available at \url{https://github.com/Gabrielcb/Unsupervised_selfAdaptative_ReID}.

After training, we remove the last classification layer from all backbones and use the last layer's output as our feature embedding. We trained our pipeline using the three backbones independently in all scenarios of adaptation. Considering the flows depicted in Figure~\ref{fig:overview_pipeline}, we perform $K_1=50$ cycles of the blue flow (50 iterations), and, in each one, we perform $K_2=5$ cycles of the orange flow (5 epochs).  We consider Adam~\cite{kingma2014adam} as the network optimizer and set the learning rate to $0.0001$ in the first 30 iterations. After the $30^{th}$ iteration, we divided it by ten and kept it unchanged until reaching the maximum number of iterations. As we show in our experiments, we can set the weight decay to zero since our proposed Cross-Camera Triplet Creation can regularize the model without extra hyper-parameters. The triplet batch size is set to 30; batches with 30 triplets are used to update the model in each epoch. The margin in Equation~\ref{eq:LWHTL} is set to $0.3$, and the number of anchors is set to $m = 2$. We resize the images to $256 \times 128 \times 3$ and apply Random Flipping and Random Erasing as data augmentation strategies during training.

\subsection{Comparison with the Prior Art}
Tables~\ref{tab:MarketAndDuke} and~\ref{tab:msmt17} show results comparing the proposed method to the state of the art. The proposed method outperforms the other methods regarding mAP and Rank-1 in Market $\rightarrow $ Duke by improving those values in 1.8 and 1.7 percentage points (p.p.), respectively, and without re-ranking. In the Duke $\rightarrow $ Market scenario, we obtain a solid competitive performance by having values 0.1 p.p. lower only in Rank-1, also without re-ranking. 
 
In turn, ABMT applies k-reciprocal encoding during training, which is more robust than Euclidean distance. 
However, it is more expensive to calculate as it is necessary to search for k-reciprocal neighbors of each feature vector in each iteration of the algorithm before clustering. In our case, we only apply the standard Euclidean distance during training, reducing the training time and complexity on adaptation, but still obtaining performance gains. Moreover, we have a single-term and single-hyper-parameter loss function, while ABMT depends on a loss with three terms and more hyper-parameters. They apply a teacher-student strategy to their training while we perform ensembling only for inference. Therefore, with a more direct pipeline and ensemble prediction, the proposed method has a Rank-1 only 0.1 p.p. lower in the Duke $\rightarrow $ Market, while outperforming all methods in all other adaptation scenarios. 

%We argue it happens since the authors adopt an ensemble-based method leveraging three deep models in a co-teaching training regime with a four-term loss function with three hyper-parameters (as we will explain later on Section \ref{sec:discussion}). The several hyper-parameters calibration may lead to an over approximation of the query and its true matches in the test set. Our work keeps three independent models without the need of co-supervision to avoid adding complexity and leveraging a simpler loss function with only one hyper-parameter. Considering the closest true gallery match image to the query (R1), our ensemble is able to retrieve more correct matches as shown on Table \ref{tab:MarketAndDuke} where our method outperform them in 1.2 p.p. in Rank-1 without re-ranking. However, SSKD co-teaching enables them to get the majority of true gallery images closer to the query image than our work, which results in a higher R5 and R10.  Note that even with a lot less  hyper-parameters in comparison to SSKD and with a simpler training process (no co-teaching, simpler loss function and late ensembling), our method scores only 0.3 p.p. less on R5 and 0.4 p.p. less on R10, showing competitive performance considering the training complexity trade-off.}

However, to benefit from the k-reciprocal encoding, we also apply it during inference to keep a simpler training process. In this case, the proposed method outperforms the methods in the prior art regarding mAP and Rank-1 in all adaptation scenarios.  \par 

%\textcolor{red}{%(R2C5) - 
Compared to SSKD in Duke $\rightarrow$ Market scenario, we are below it by 0.3 and 0.4 p.p. in Rank-5 and Rank-10, respectively. Considering the closest actual gallery match image to the query (R1), our ensemble retrieves more correct matches, as Table~\ref{tab:MarketAndDuke} shows, with our method outperforming SSKD by 1.2 p.p. in Rank-1 without re-ranking. Even with fewer hyper-parameters than SSKD and a more straightforward training process (no co-teaching, simpler loss function, and late ensembling), our method shows competitive results considering the training complexity trade-off.%}

Interestingly, the proposed method performs better under more difficult adaptation scenarios. We measure the difficulty of a scenario based on the number of different cameras it comprises. Market, Duke, and MSMT17 have 6, 8, and 15 cameras, respectively. Hence the most challenging adaptation scenario is from Market to MSMT17. We adapt a model from a simpler scenario (6 cameras, all videos recorded in the same day period and the same season of the year) to a more complex target domain (15 cameras -- 12 outdoors and 3 indoors -- recorded at 3 different day periods -- morning, afternoon and noon -- in 4 different days -- each day on a different season of the year). Market $ \rightarrow $ MSMT17 is the most challenging adaptation and close to real-world conditions where we might have people recorded along the day and in different locations (indoors and outdoors). In this case, as shown in Table~\ref{tab:msmt17}, we obtained the highest performance even without re-ranking techniques. The proposed method outperforms the state of the art by 1.5 and 2.1 p.p. in mAP and Rank-1, respectively, on Duke $\rightarrow$ MSMT17, and by 2.2 and 4.2 p.p. on the most challenge scenario, Market $\rightarrow$ MSMT17.

There are several reasons why our method performs well. We explicitly design a model to deal with the diversity of cameras and viewpoints by creating a set of triplets based on the different cameras in a cluster. We also keep a more straightforward training, with only one hyper-parameter in our loss function (triplet loss margin). Most works in the ReID literature optimize a loss function with many terms and hyper-parameters. They usually consider the Duke $\rightarrow$ Market or the Market $\rightarrow $ Duke scenarios (or both of them) to perform grid-searching over hyper-parameter values. Once they find the best values, they keep them unchanged for all adaptation setups. 

In ABMT~\cite{chen2020enhancing}, the authors do not provide a clear explanation on how they define the hyper-parameter values for their loss function. However, they perform an ablation study over Duke $\rightarrow$ Market and Market $\rightarrow $ Duke scenarios, so their results might be biased to those specific setups, which gives them one of the best performances. However, when they keep the same values for different and more challenging scenarios, such as Market $\rightarrow $ MSMT17 or Duke $\rightarrow$ MSMT17, they obtain worse results than ours by a large margin. This shows that our method provides a better generalization capability brought by a simpler loss function and a more diverse training. It prevents us from choosing specific hyper-parameter values and be biased to a specific adaptation setup. Consequently, we achieve the best performances, especially in the most challenging scenarios. 

\begin{table*}[!ht]
\caption{Results on Market1501 to DukeMTMC-ReID and DukeMTMCRe-ID to Market1501 adaptation scenarios. We report mAP, Rank-1, Rank-5, and Rank-10, comparing to several state-of-art methods. The best result is shown in \textbf{bold}, the second in \underline{underline} and the third in \textit{italic}. Works with (*) do not pre-train the model in any source dataset before adaptation.}
\label{tab:MarketAndDuke}
\centering
\begin{tabular}{|p{2.7cm}| p{2.0cm}|p{0.8cm}|p{1.0cm}|p{1.0cm}|p{1.0cm}|p{0.8cm}|p{1.0cm}|p{1.0cm}|p{1.0cm}|}
\hline
\multicolumn{1}{|c|}{} &
\multicolumn{1}{|c|}{} &
\multicolumn{4}{|c|}{Duke $\rightarrow$ Market} & \multicolumn{4}{|c|}{Market $\rightarrow$ Duke} \\
\hline
Method & reference & mAP & R1 & R5 & R10
& mAP & R1 & R5 & R10 \\ \hline
SSL \cite{lin2020unsupervised}* & CVPR 2020 & 37.8 & 71.7 & 83.8 & 87.4 & 28.6 & 52.5 & 63.5 & 68.9 \\
%\textcolor{red}{CCSE \cite{lin2020unsupervisedccse}*} & \textcolor{red}{TIP 2020} & \textcolor{red}{38.0} & \textcolor{red}{73.7} & \textcolor{red}{84.0} & \textcolor{red}{87.9} & \textcolor{red}{30.6}& \textcolor{red}{56.1} & \textcolor{red}{66.7} & \textcolor{red}{71.5} \\ 
CCSE \cite{lin2020unsupervisedccse}* & TIP 2020 & 38.0 & 73.7 & 84.0 & 87.9 & 30.6 & 56.1 & 66.7 & 71.5 \\
UDAP \cite{song2020unsupervised} & PR 2020 & 53.7 & 75.8 & 89.5 & 93.2 & 49.0 & 68.4 & 80.1 & 83.5 \\
MMCL \cite{wang2020unsupervised}& CVPR 2020 & 60.4 & 84.4 & 92.8 & 95.0 & 51.4 & 72.4 & 82.9 & 85.0 \\
ACT \cite{yang2020asymmetric} & AAAI 2020 & 60.6 & 80.5 & - & - & 54.5 & 72.4 & - & - \\
ECN-GPP \cite{zhong2020learning} & TPAMI 2020 & 63.8 & 84.1 & 92.8 & 95.4 & 54.4 & 74.0 & 83.7 & 87.4 \\
HCT \cite{zeng2020hierarchical}* & CVPR 2020 & 56.4 & 80.0 & 91.6 & 95.2 & 50.7 & 69.6 & 83.4 & 87.4 \\ 
SNR \cite{jin2020style} & CVPR 2020 & 61.7 & 82.8 & - & - & 58.1 & 76.3 & - & - \\
AD-Cluster \cite{zhai2020ad} & CVPR 2020 & 68.3 & 86.7 & 94.4 & 96.5 & 54.1 & 72.6 & 82.5 & 85.5 \\
MMT \cite{ge2020mutual} & ICLR 2020 & 71.2 & 87.7 & 94.9 & 96.9 & 65.1 & 78.0 & 88.8 & \textit{92.5} \\
CycAs \cite{wang2020cycas}* & ECCV 2020 & 64.8 & 84.8 & - & - & 60.1 & 77.9 & - & - \\
DG-Net++ \cite{zou2020joint} & ECCV 2020 & 61.7 & 82.1 & 90.2 & 92.7 & 63.8 & 78.9 & 87.8 & 90.4 \\
MEB-Net \cite{zhai2020multiple} & ECCV 2020 & 76.0 & 89.9 & 96.0 & 97.5 & 66.1 & 79.6 & 88.3 & 92.2 \\
Dual-Refinement \cite{dai2020dual} & arXiv 2020 & 78.0 & 90.9 & \textit{96.4} & \textit{97.7} & 67.7 & 82.1 & 90.1 & \textit{92.5} \\
SSKD \cite{yin2020sskd} & arXiv 2020 & \textit{78.7} & 91.7 & \textbf{97.2} & \textbf{98.2} & 67.2 & 80.2 & \textit{90.6} & \underline{93.3} \\ 
ABMT \cite{chen2020enhancing} & WACV 2020 & \underline{80.4} & \underline{93.0} & - & - & \textit{70.8} & \textit{83.3} & - & - \\
\hline
%\textcolor{red}{\textbf{Ours (w/o Re-Ranking)*}} & \textcolor{red}{This Work} & \textcolor{red}{67.7} & \textcolor{red}{89.5}  & \textcolor{red}{94.8} & \textcolor{red}{96.5} & \textcolor{red}{68.8} & \textcolor{red}{82.4} & \textcolor{red}{\textit{90.6}} & \textcolor{red}{\textit{92.5}} \\ \hline
\textbf{Ours (w/o Re-Ranking)*} & This Work & 67.7 & 89.5  & 94.8 & 96.5 & 68.8 & 82.4 & \textit{90.6} & \textit{92.5} \\ \hline
\textbf{Ours (w/o Re-Ranking)} & This Work & 78.4 & \textit{92.9} & \underline{96.9} & \underline{97.8} & \underline{72.6} & \underline{85.0} & \underline{92.1} & \textbf{93.9} \\ \hline
\textbf{Ours (w/ Re-Ranking)} & This Work & \textbf{88.0} & \textbf{93.8} & \textit{96.4} & 97.4 & \textbf{82.7} & \textbf{87.2} & \textbf{92.5} & \textbf{93.9} \\ \hline
\end{tabular}
\end{table*}

\subsection{Discussion}
\label{sec:discussion}

As we aim to re-identify people in a camera system in an unsupervised way, we must be robust to hyper-parameters that require adjustments based on grid-searching using true label information, keeping the training process (and adaptation to a target domain) as simple as possible. If a pipeline is complex and too sensitive to hyper-parameters, it might be challenging to train and deploy it on a real investigation scenario, where we do not have prior knowledge about the people of interest. This complexity leads to sub-optimal performance. 
%In this sense, algorithms should be less sensitive to hyper-parameter or ideally reduce them on adaptation stage. 
This has already been pointed out in~\cite{dubourvieux2020unsupervised}. The authors claim that most works rely on many hyper-parameters during the adaptation stage, which can help or hinder the performance, depending on the value assigned to them and which adaptation scenario is considered.

SSKD\cite{yin2020sskd} is an ensemble-based method leveraging three deep models in a co-teaching training regime with a four-term loss function with three hyper-parameters. One of the terms of their final loss function is a multi-similarity loss~\cite{wang2019multi}, with three extra hyper-parameters to train the model.  

MEB-Net has complex training by relying on a co-training technique with three deep neural networks in which each one learns with the others. Each of these three networks has its separate loss function with six terms, and their overall loss function is a weighted average of the individual loss functions from each model on the ensemble. 

ABMT also leverages a teacher-student model where the teacher and student networks share the same architecture, increasing time and memory complexity during training. Moreover, they utilize a three-term loss function to optimize both models with three hyper-parameters controlling the contribution of each term to the final loss. They update the teacher weights based on the exponential moving average (EMA) of the student weights, in order to avoid error label amplification on training. This also adds another parameter to control the inertia in the teacher weights' EMA. The authors do not perform an ablation study regarding the hyper-parameter value variation to assess their impact on final performance. %In turn, we perform self-ensembling in our work by weighing each checkpoint's contribution directly considering the amount of target data used in the iteration. This prevents the need of additional hyper-parameters.\par 

Based on these observations, our proposed model better captures the diversity of real cases, by considering a loss function with a single term and that is less sensitive to hyper-parameters (only margin $\alpha$ needs to be selected). In such setups, it is difficult to select hyper-parameter values correctly, as we might not know any information about the identities on the target domain. The self-ensembling also summarizes the whole training into a single model by using each checkpoint's confidence values over the target data, without using any hyper-parameter or human-defined value. Even adopting a more straightforward formulation, we still obtain state-of-the-art performance on the Market $ \rightarrow $ Duke scenario and competitive performance on the Duke $\rightarrow$ Market scenario. Each architecture in our work is trained in parallel without any co-teaching strategy. After self-ensembling, the joint contribution from different backbones is applied only on evaluation time, avoiding label propagation of noisy examples (e.g., potential outliers) but still taking advantage of the complementarity between them.

%\textcolor{red}{%(R1C1) 
Our assumptions are the same as recent prior art~\cite{zhai2020ad, zhong2020learning, lin2020unsupervisedccse}. We assume to know from which camera an image of a person was recorded but not the identity. We rely on camera information to filter out clusters elements captured by only one camera and create the cross-camera triplets.%} 

%\textcolor{red}{
We also assume that at least two cameras have captured most identities and all of them have non-overlapping vantage points. All prior art holds this assumption as defined by the datasets and train/test split division.%}

%\textcolor{red}{
Finally, we assume that training on a source domain related to Person Re-Identification gives the model a basic knowledge to adapt to the target domain. This knowledge enables the model to propose better initial clusters on early iterations, grouping feature vectors from the same identity recorded from different cameras. The pipeline starts the adaptation with more reliable pseudo-labels in the clustering step and progressively creates more clusters representing more identities on the target domain. All works hold this assumption in Table~\ref{tab:MarketAndDuke} that do not have the (*) after their name. %}

%\textcolor{red}{
Section~\ref{sec:ablation_study} shows that our pipeline still performs well even without pre-training in a source dataset. In other words, we take the backbone trained over ImageNet and directly apply it without any previous ReID-related knowledge. Even in this setup, we can achieve competitive performance.%}

\begin{table*}[ht]
\caption{Results on Market1501 to MSMT17 and DukeMTMCRe-ID to MSMT17 adaptation scenarios. We report mAP, Rank-1, Rank-5, and Rank-10, comparing to several state-of-art methods. The best result is shown in \textbf{bold}, the second in \underline{underline} and the third in \textit{italic}. Works with (*) do not pre-train the model in any source dataset before adaptation.}
\label{tab:msmt17}
\centering
\begin{tabular}{|p{2.7cm}| p{2.0cm}|p{0.8cm}|p{1.0cm}|p{1.0cm}|p{1.0cm}|p{0.8cm}|p{1.0cm}|p{1.0cm}|p{1.0cm}|}
\hline
\multicolumn{1}{|c|}{} &
\multicolumn{1}{|c|}{} &
\multicolumn{4}{|c|}{Duke $\rightarrow$ MSMT17} & \multicolumn{4}{|c|}{Market $\rightarrow$ MSMT17} \\
\hline
Method & reference & mAP & R1 & R5 & R10
& mAP & R1 & R5 & R10 \\ \hline
%PTGAN \cite{wei2018person} & CVPR 2018 & 3.3 & 11.8 & - & 27.4 & 2.9 & 10.2 & - & 24.4 \\
ECN \cite{zhong2019invariance} & CVPR 2019 & 10.2 & 30.2 & 41.5 & 46.8 & 8.5 & 25.3 & 36.3 & 42.1 \\
%\textcolor{red}{CCSE \cite{lin2020unsupervisedccse}*} & \textcolor{red}{TIP 2020} & \textcolor{red}{9.9} & \textcolor{red}{31.4} & \textcolor{red}{41.4} & \textcolor{red}{45.7} & \textcolor{red}{9.9} & \textcolor{red}{31.4} & \textcolor{red}{41.4} & \textcolor{red}{45.7} \\
%\textcolor{red}{SSG \cite{fu2019self}} & \textcolor{red}{ICCV 2019} & \textcolor{red}{13.3} & \textcolor{red}{32.2} & \textcolor{red}{-} & \textcolor{red}{51.2} & \textcolor{red}{13.2} & \textcolor{red}{31.6} & \textcolor{red}{-} & \textcolor{red}{49.6} \\
CCSE \cite{lin2020unsupervisedccse}* & TIP 2020 & 9.9 & 31.4 & 41.4 & 45.7 & 9.9 & 31.4 & 41.4 & 45.7 \\
SSG \cite{fu2019self} & ICCV 2019 & 13.3 & 32.2 & - & 51.2 & 13.2 & 31.6 & - & 49.6 \\
ECN-GPP \cite{zhong2020learning} & TPAMI 2020 & 16.0 & 42.5 & 55.9 & 61.5 & 15.2 & 40.4 & 53.1 & 58.7 \\
MMCL \cite{wang2020unsupervised} & CVPR 2020 & 16.2 & 43.6 & 54.3 & 58.9 & 15.1 & 40.8 & 51.8 & 56.7 \\
MMT \cite{ge2020mutual} & ICLR 2020 & 23.3 & 50.1 & 63.9 & 69.8 & 22.9 & 49.2 & 63.1 & 68.8 \\
CycAs \cite{wang2020cycas}* & ECCV 2020 & 26.7 & 50.1 & - & - & 26.7 & 50.1 & - & - \\
DG-Net++  \cite{zou2020joint} & ECCV 2020 & 22.1 & 48.8 & 60.9 & 65.9 & 22.1 & 48.4 & 60.9 & 66.1 \\
Dual-Refinement \cite{dai2020dual} & arXiv 2020 & 26.9 & 55.0 & 68.4 & 73.2 & 25.1 & 53.3 & 66.1 & 71.5 \\
SSKD \cite{yin2020sskd} & arXiv 2020 & 26.0 & 53.8 & \textit{66.6} & \textit{72.0} & 23.8 & 49.6 & 63.1 & 68.8 \\ 
ABMT \cite{chen2020enhancing} & WACV 2020 & \textit{33.0} & \textit{61.8} & - & - & 27.8 & 55.5 & - & - \\ 
SpCL \cite{ge2020self} & NeurIPS 2020 & - & - & - & - & \textit{31.0} & \textit{58.1} & \textit{69.6} & \textit{74.1} \\ \hline
\textbf{Ours (w/o Re-Ranking)} & This Work & \underline{34.5} & \underline{63.9} & \underline{75.3} & \underline{79.6} & \underline{33.2} & \underline{62.3} & \underline{74.1} & \underline{78.5} \\ \hline
\textbf{Ours (w/ Re-Ranking)} & This Work & \textbf{46.6} & \textbf{69.6} & \textbf{77.1} & \textbf{80.4} & \textbf{45.2} & \textbf{68.1} & \textbf{76.0} & \textbf{79.2} \\ \hline
\end{tabular}
\end{table*}

%\textcolor{red}{
\subsection{Qualitative Analysis}%}%(R1C3)

\begin{figure*}[ht]
\centering
\subfloat[]{\includegraphics[width=3.5in]{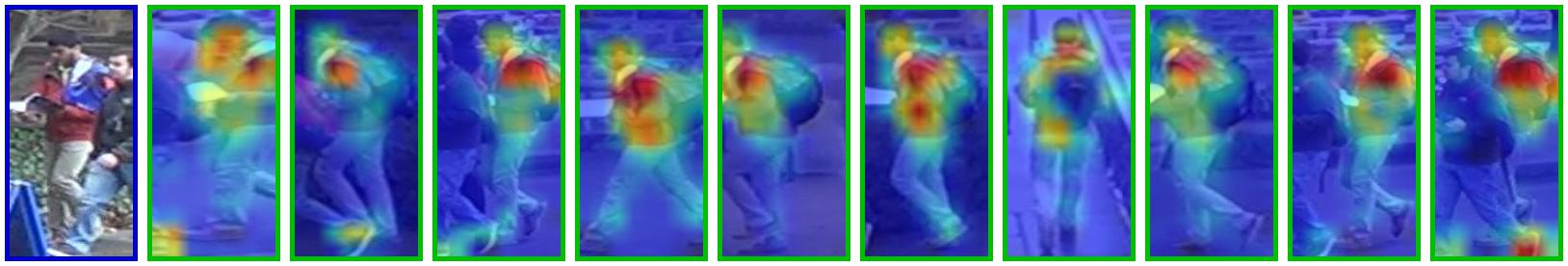}
\label{fig:M2D_successful}}
\hfil
\subfloat[]{\includegraphics[width=3.5in]{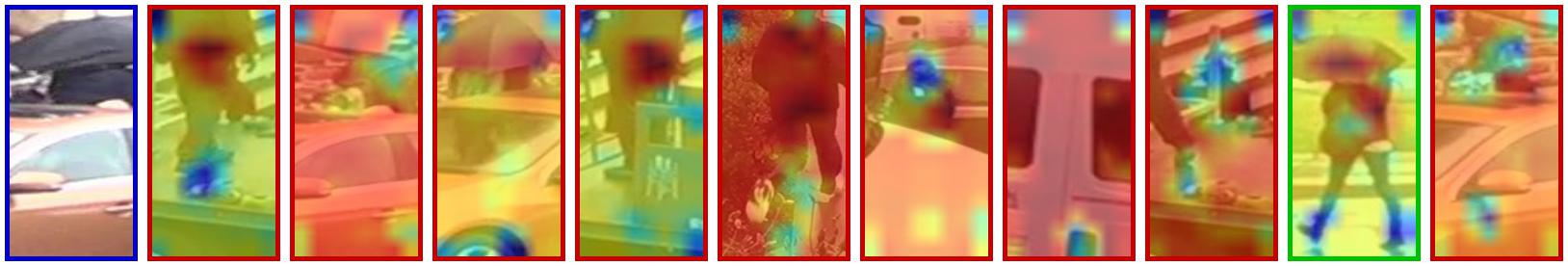} 
\label{fig:M2D_failure}}
\caption{The most activated regions in the gallery image given a query on DukeMTMCReID (Market $\rightarrow$ Duke scenario) for ResNet50. (a) Successful match; (b) Failure case.}
\label{fig:QualitativeM2D}
\end{figure*}

\begin{figure*}[ht]
\centering
\subfloat[]{\includegraphics[width=3.5in]{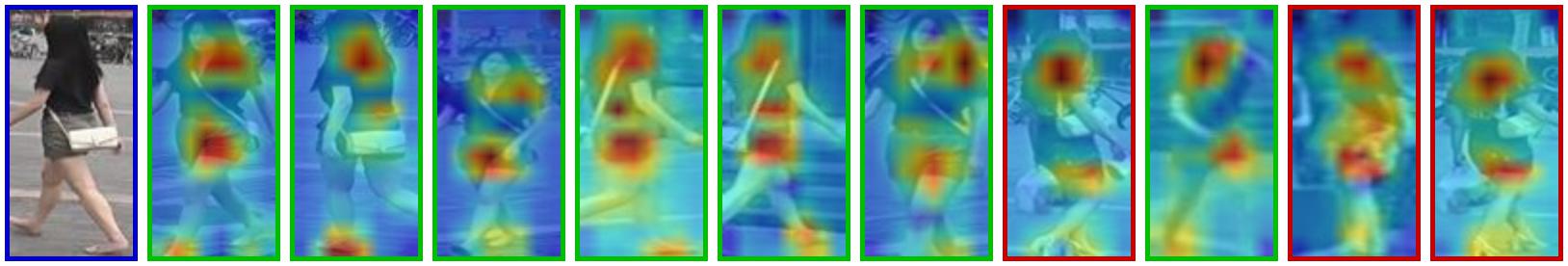}
\label{fig:D2M_successful}}
\subfloat[]{\includegraphics[width=3.5in]{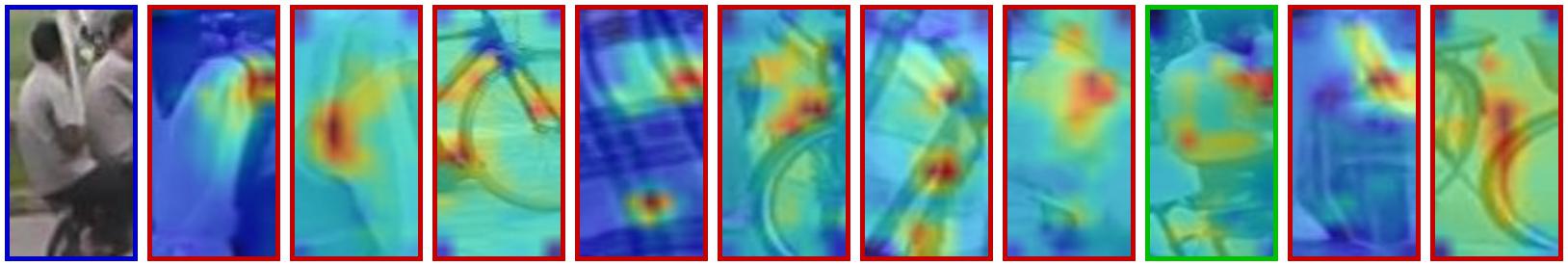}
\label{fig:D2M_failure}}
\caption{Highlighting image regions most activated on gallery image given query on Market1501 after run Duke $\rightarrow$ Market scenario on ResNet50. (a) Successful match; (b) Failure case.}
\label{fig:QualitativeD2M}
\end{figure*}

%\textcolor{red}{
We now provide qualitative analysis by highlighting regions of the top-10 gallery images returned for a given query image. The redder the color of a region, the more important it is to the ranking. As explained in Section~\ref{subsec:datasets}, the correct matches always come from cameras different from the query’s camera. The green contour denotes a true positive, the red contour a false positive, and the blue color the query image. We present successful cases (when the first gallery image is a true positive) and failure cases (when the first gallery image is a false positive) for each camera on Market1501 and DukeMTMC-ReID datasets. We show two successful cases and two failure cases (one for each dataset) in Figures~\ref{fig:QualitativeM2D} and~\ref{fig:QualitativeD2M} considering ResNet50 as the backbone. For visualizations for all cameras of both datasets, please refer to the Supplementary Material. MSMT17 was not considered as the dataset agreement does not allow the reproduction of the images in any format.%}

%\textcolor{red}{
Figures~\ref{fig:M2D_successful} and~\ref{fig:D2M_successful} depict two successful cases on Market $\rightarrow$ Duke and Duke $\rightarrow$ Market scenarios, respectively. In both cases, we see that our model finds fine-grained details on the image leading to a correct match.  As an example, Figure~\ref{fig:M2D_successful} shows the model focusing on the red jacket, even in a different pose and under occlusion ($7^{th}$ and $10^{th}$ image from left to right). Figure~\ref{fig:D2M_successful} shows that the model can overcome pose changes of the query on a cross-view setup. The query only shows the person's back, but the closest image is a true match showing the person from the front. The same happens on the second closest image, where the identity has its back recorded by another camera; and on the fourth and fifth closest images, only the right side is captured. The third closest image not only records a different position of the query, but also has a different resolution. This shows that the model effectively overcomes identity pose changes and resolution on cross-view cameras. %}

%\textcolor{red}{
Figures~\ref{fig:M2D_failure} and~\ref{fig:D2M_failure} depict failure cases to show the limitations of the method. The errors happen when there is no person on the image --- see \ref{fig:M2D_failure}, which has been fully occluded by the car. In this case, the method does not have any specific region to focus on, and then the gallery images are almost fully activated. Another failure case happens when the identity is on a motorcycle (Figure \ref{fig:D2M_failure}) along with another identity which led to mismatching cases where there is no identity (distractor images on the gallery) or images with parts of a bike. In the  Supplementary Material, we provide more successful and failure cases in other cameras for both datasets.%} 

%(R3C1) 
%\textcolor{red}{
\subsection{Results on an Unsupervised Scenario} %}
\label{subsec:results_unsupervised}

%\textcolor{red}{
This section explores the possibilities of our method when not performing any pre-training on a source domain. Here the method starts with backbones trained over ImageNet directly. This is a harder case as we eliminate the possibility of having prior knowledge of the person re-identification problem. This requires the backbones to adapt themselves to the target, not relying on any identity-related annotation coming from the source domain. Table~\ref{tab:MarketAndDuke} shows the results denoted by ``Ours(w/o Re-Ranking)*''. In this case, we keep $\xi = 0.05$ when Duke is the target, as in previous results, and $\xi = 0.03 $ when Market is the target. The value $\xi = 0.05 $ was too strict, leading to clusters with images from only one camera for the Market dataset. Section~\ref{sec:ablation_study} presents a deeper analysis of different choices of $\xi$ on the clustering process. %}

%\textcolor{red}{All models in Table~\ref{tab:MarketAndDuke} assume source pre-training except the ones denoted with the star. Our model obtains competitive performance even without pre-training and without re-ranking. When Market is the target domain, we lose 10.7 and 3.4 p.p. on mAP and Rank-1 compared to our equivalent model considering the source domain. Compared to ABMT, we lose 12.7 and 3.5 p.p. on mAP and Rank-1, respectively. This shows that the model becomes more sensitive to retrieve cross-camera true matches, since they are farther apart (lower mAP) compared to the equivalent pre-trained model and ABMT.%} 

%\textcolor{red}{
However, when we consider Duke as the target domain, the model without source pre-training is the third best. We lose 3.8 and 2.6 p.p. to the equivalent pre-trained model in mAP and Rank-1, respectively, and we lose 2.0 and 0.9 p.p. compared to ABMT, outperforming all other methods. This shows that, although our model is not completely robust to the backbone initialization, it is still capable of mining discriminative features, even without pre-training, proving comparative or better results when compared to the state of the art.%}

%\textcolor{red}{
The proposed method outperforms all others in the same conditions (no pre-training, denoted with a star in Table~\ref{tab:MarketAndDuke}). The difference to the best one (CycAs) is 2.9 and 4.7 p.p. on mAP and Rank-1 when Market is the target, and in 8.7 and 4.5 p.p. on mAP and Rank-1 when Duke is the target. %}

%\textcolor{red}{
We conclude that the previous training on a ReID source-related dataset is important for better performances on the task. However, when no ReID source domain is available, our methods can still provide competitive results, mainly in the more challenging scenario (Duke as target). %} 

%\begin{table*}[ht]
%\caption{Impact when we \textbf{do not} apply pre-training and leverage each backbone trained over ImageNet directly on the pipeline. Each individual results was obtained after self-Ensembling. Best values are in \textbf{bold}.}
%\label{tab:ensemble_unsupervised}
%\centering
%\begin{tabular}{|P{2.5cm}|P{0.8cm}|P{1.0cm}|P{1.0cm}|P{1.0cm}|P{0.8cm}|P{1.0cm}|P{1.0cm}|P{1.0cm}|}
%\hline
%\multicolumn{1}{|c|}{} &
%\multicolumn{4}{|c|}{Market} & \multicolumn{4}{|c|}{Duke} \\
%\hline
%& mAP & R1 & R5 & R10 & mAP & R1 & R5 & R10 \\ \hline
%ResNet & 60.3 & 83.9 & 92.1 & 94.5 & 62.7 & 78.0 & 87.6 & 90.7 \\
%OSNet & 59.0 & 85.0 & 92.4 & 94.3 & 62.5 & 78.5 & 87.7 & 90.3 \\
%DenseNet & 62.0 & 86.8 & 93.8 & 95.3 & 64.7 & 79.9 & 88.7 & 91.4 \\
%Ensembled model  & \textbf{67.7} & \textbf{89.5} & \textbf{94.8} & \textbf{96.5} & \textbf{68.8} & \textbf{82.4} & \textbf{90.6} & \textbf{92.5} \\
%\hline
%\end{tabular}
%\end{table*}

\section{Ablation Study}
\label{sec:ablation_study}
This section shows the contribution of each part of the pipeline to the final result. In each experiment, we change one of the parts and keep the others unchanged. If not explicitly mentioned, we consider ResNet50 as the backbone, OPTICS with hyper-parameter $ \xi = 0.05 $, and self-ensembling applied after training.  

\subsection{Impact of the Clustering Hyper-parameter}
\label{subsec:impact_of_xi}

\begin{figure}[ht]
\centering
\subfloat[]{\includegraphics[width=1.6in]{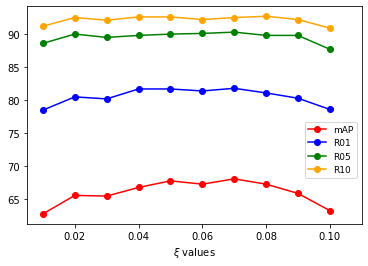}
\label{fig:xi_market2duke}}
\hfil
\subfloat[]{\includegraphics[width=1.6in]{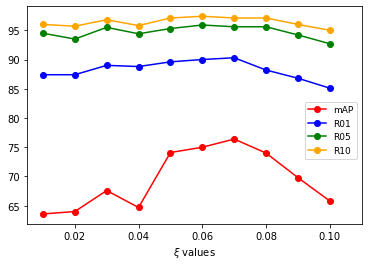}
\label{fig:xi_duke2market}}
\caption{Impact of clustering hyper-parameter $ \xi $. Results on (a) Market $ \rightarrow $ Duke, and (b) Duke $ \rightarrow $ Market.}
\label{fig:ablation_study_xi}
\end{figure}

Although we have only one hyper-parameter in the loss function, we still need to set hyper-parameter $ \xi $  of the OPTICS clustering algorithm, a threshold in the range $[0,1]$. The closer $ \xi $ is to 1, the stronger is the criteria to define a cluster; that is, we might have many samples not assigned to any cluster, which leads to several detected outliers (if $ \xi = 1$, all feature vectors are detected as outliers). In contrast, the closer $ \xi $ is to 0, the more relaxed the criteria is, and more samples are assigned to clusters (if $ \xi = 0 $, all feature vectors are grouped into a single cluster). In Figure~\ref{fig:ablation_study_xi}, we show the impact of the threshold $ \xi $ for the Market $ \rightarrow $ Duke and Duke $ \rightarrow $ Market scenarios.

% In our experiments, we observe that if $ \xi \geq 0.2 $ many feature vectors are still detected as outliers on Market $\rightarrow$ Duke scenario. Hence, we performed experiments changing the $ \xi $ value as Figure~\ref{fig:ablation_study_xi} shows. 
%

%(R2C4) - 
%\textcolor{red}{
The best value for $ \xi $ changes according to the adaptation scenario. This is expected when dealing with different unseen target domains. In both cases, Rank-1, Rank-5, and Rank-10 curves are more stable than the mAP curve, showing that the parameter does not impact the retrieval of true positive images. The best Rank-1 values are obtained for $ \xi $ between $0.04$ and $0.08$ considering both scenarios and, in the more challenging one (Market $ \rightarrow $ Duke), it achieves the second-best value when $ \xi = 0.05$, for both mAP and Rank-1. Although the best performance is achieved when $\xi = 0.07$ (best mAP and Rank-1), it relies on an unstable point in the setup of Duke $\rightarrow$ Market, and it is only marginally better than $\xi = 0.05 $ for Market $\rightarrow $ Duke. Rank-5 and Rank-10 tend to be more stable in both cases. Thus we adopt $ \xi = 0.05 $ in all scenarios.%} 

\subsection{Impact of Curriculum Learning}

In our pipeline, Stage 3 is responsible for cluster selection. After running the clustering algorithm, a feature vector can be an outlier, assigned to a cluster with only one camera or assigned to a cluster with two or more cameras. We argue that feature space cleaning is essential for better adaptation, and that feature vectors in a cluster with at least two cameras are more reliable than ones assigned as outliers or to cluster with a single camera. Then, we consider the curriculum learning principle to select the most confident samples and learn in an easy-to-hard manner. To achieve this, we remove the outliers and the clusters with only one camera. To check the impact of this removal, we performed four experiments in which we alternate between keeping the outliers and the clusters with only one camera. The results are summarized in Table~\ref{tab:ablation_cluster_selection}.

\begin{table*}[ht]
\caption{Impact of curriculum learning, when considering different cluster selection criteria. We tested our method with and without outliers and with and without clusters with only one camera in the feature space. All experiments consider ResNet50 as the backbone with self-ensembling applied after training.}
\label{tab:ablation_cluster_selection}
\centering
\begin{tabular}{|P{1.5cm}| P{2.0cm}|P{0.8cm}|P{1.0cm}|P{1.0cm}|P{1.0cm}|P{0.8cm}|P{1.0cm}|P{1.0cm}|P{1.0cm}|}
\hline
\multicolumn{1}{|c|}{} &
\multicolumn{1}{|c|}{} &
\multicolumn{4}{|c|}{Duke $\rightarrow$ Market} & \multicolumn{4}{|c|}{Market $\rightarrow$ Duke} \\
\hline
w/o outliers & w/o cluster with one camera & mAP & R1 & R5 & R10 & mAP & R1 & R5 & R10 \\ \hline
- & - & 50.9 & 79.2 & 89.5 & 92.8 & 32.7 & 56.7 & 68.5 & 72.9 \\
\checkmark & - & 72.4 & 89.5 & 95.2 & 96.7 & 66.8 & 81.1 & \textbf{90.2} & 92.4 \\
 - & \checkmark & 49.1 & 79.8 & 89.5 & 92.6 & 32.7 & 57.2 & 68.4 & 72.3 \\
\checkmark & \checkmark & \textbf{74.1} & \textbf{89.6} & \textbf{95.3} & \textbf{97.1} & \textbf{67.8} & \textbf{81.7} & 90.0 & \textbf{92.6} \\\hline
\end{tabular}
\end{table*}

We observe a performance gain on most metrics, especially on mAP and Rank-1, when we apply our cluster selection strategy. If we keep the outliers in the feature space (first and third rows in Table~\ref{tab:ablation_cluster_selection}), we face the most significant performance drop in both adaptation scenarios. It shows the importance of removing outliers after the clustering stage; otherwise, they can be considered in the creation of triplets, increasing the number of false negatives (for instance, selecting negative samples of the same real class) and, consequently, hindering the performance. We see a lower performance drop by keeping clusters with only one camera but without outliers (second row), indicating that those clusters do not hinder the performance much, but might contain noisy samples for model updating. It is more evident when we verify that the most gains were over mAP and lower gains over Rank-1 in the last row. This demonstrates that if we keep one-camera clusters, the model can still retrieve most of the gallery's correct images but with lower confidence. Hence, the cluster selection criteria effectively improves our model generalization and we apply it in all adaptation scenarios. 

\begin{figure*}[ht]
\centering
\subfloat[]{\includegraphics[width=2.3in]{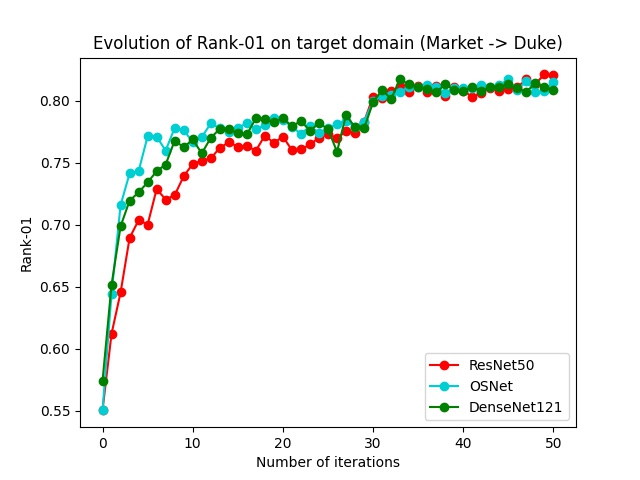}
\label{fig:rank01_market2duke}}
\hfil
\subfloat[]{\includegraphics[width=2.3in]{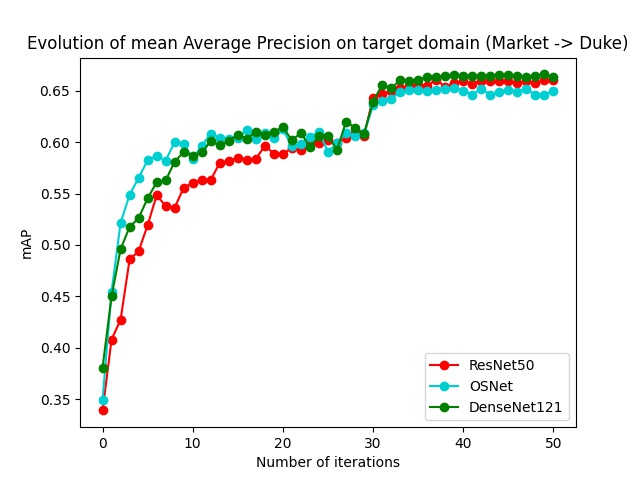}
\label{fig:mAP_market2duke}}
\hfil
\subfloat[]{\includegraphics[width=2.3in]{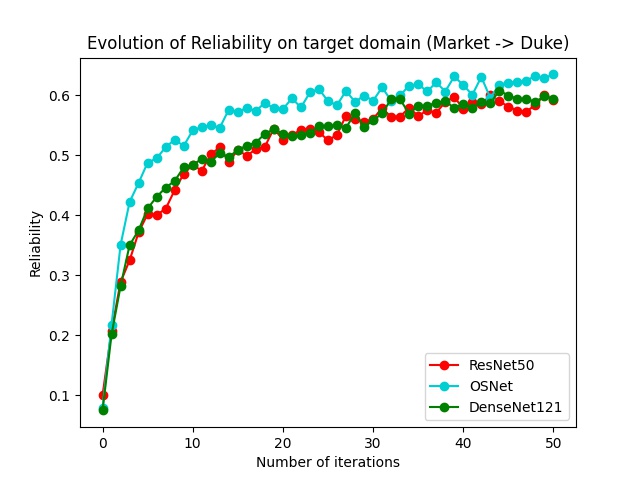}
\label{fig:reliability_market2duke}}
\caption{Progress on Rank-1, mean Average Precision and Reliability on target dataset, in the Market1501 to DukeMTMC-ReID scenario.}
\label{fig:progresses_market2duke}
\end{figure*}

\begin{figure*}[ht]
\centering
\subfloat[]{\includegraphics[width=2.3in]{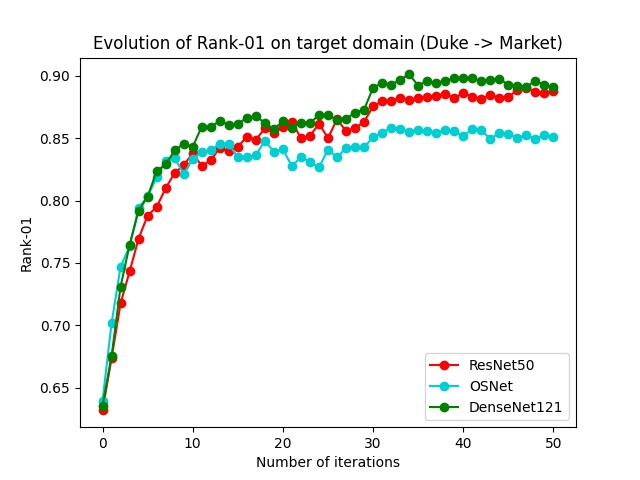}
\label{fig:rank01_duke2market}}
\hfil
\subfloat[]{\includegraphics[width=2.3in]{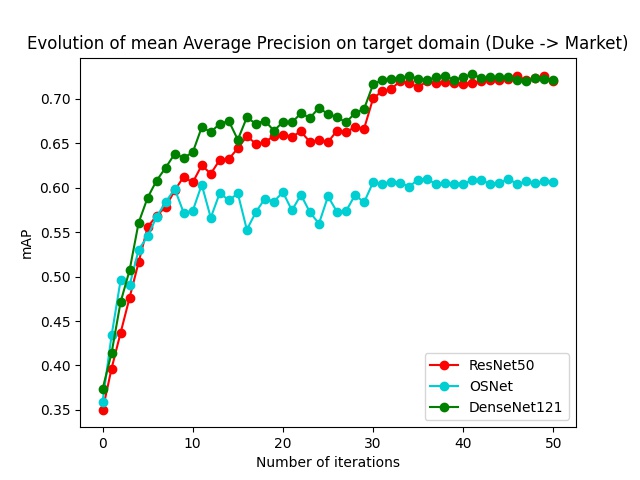}
\label{fig:mAP_duke2market}}
\hfil
\subfloat[]{\includegraphics[width=2.3in]{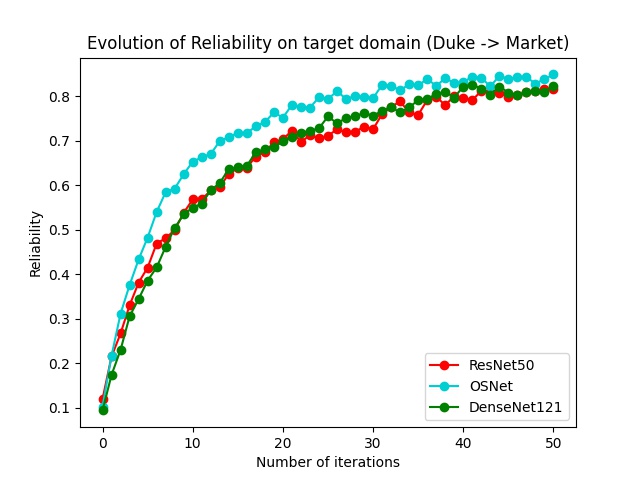}
\label{fig:reliability_duke2market}}
\caption{Progress on Rank-1, mean Average Precision and Reliability on target dataset on DukeMTMC-ReID to Market1501 scenario.}
\label{fig:progresses_duke2market}
\end{figure*}

With this strategy, we observe that the percentage of feature vectors from the target domain kept in the feature space increases during the adaptation, as shown in Figures~\ref{fig:reliability_market2duke} and~\ref{fig:reliability_duke2market}. In fact, reliability, mAP and Rank-1 increase during training (Figures~\ref{fig:progresses_market2duke} and~\ref{fig:progresses_duke2market}), which means that the model becomes more robust in the target domain as more iterations are performed. This demonstrates the curriculum learning importance, where easier examples at the beginning of the training (images whose feature vectors are assigned to clusters with at least two cameras in early iterations) are used to give an initial knowledge about the unseen target domain and allow the model to increase its performance gradually.

%(R3C3) 
% \textcolor{red}{\subsection{Removal of clusters with one camera}}

\begin{figure}[ht]
\centering
\subfloat[]{\includegraphics[width=1.65in]{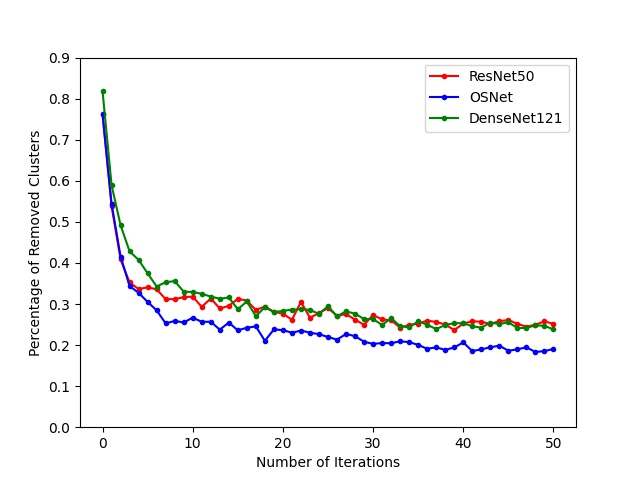}
\label{fig:cluster_remotion_market2duke}}
\hfil
\subfloat[]{\includegraphics[width=1.65in]{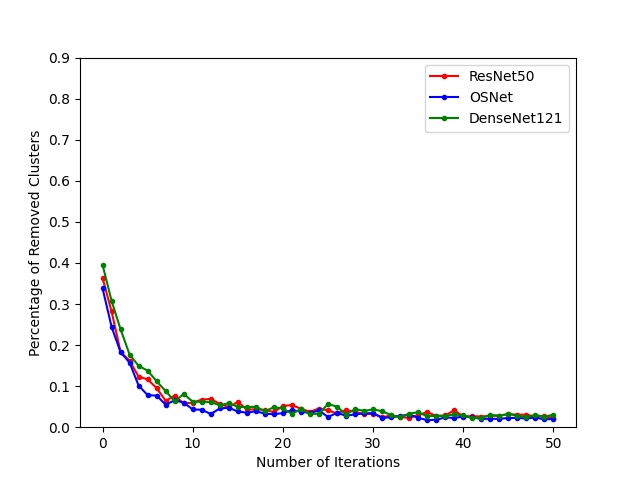}
\label{fig:cluster_remotion_duke2market}}
\caption{Percentage of cluster removed along the training iterations on (a) Market $\rightarrow$ Duke and (b) Duke $\rightarrow$ Market scenarios considering the three backbones trained independently.}
\label{fig:cluster_remotion_progress}
\end{figure}

%\textcolor{red}{
As a direct consequence, the number of clusters with only one camera removed from the feature space decreases, as shown in Figure~\ref{fig:cluster_remotion_progress}. This means that the model learns to group cross-view images in the same cluster. %}

%In Stage 3 of the proposed method  (Figure \ref{fig:overview_pipeline}), we perform the Cluster Selection based on the number of cameras in each cluster (c.f., Section \ref{subsec:cluster_selection}). As the training progresses, the model increases its knowledge of the target domain, and it can group images from cross-camera views that in past iterations were assigned as outliers or to clusters with images from only one camera. The number of clusters with images from only one camera tends to decrease as iterations increase. Consequently, the number of removed clusters also tends to decrease. Figure~\ref{fig:cluster_remotion_progress} shows this trend.} 

%\textcolor{red}{
For the Market $\rightarrow $ Duke scenario, the initial percentage of removed clusters is higher than on Duke $\rightarrow$ Market. This is expected as the former is a more complex case, so initial clusters tend to have several images grouped due to the camera bias, which leads to a higher number of clusters comprising images recorded from only one camera. For the same reason, the final percentage for Market $\rightarrow$ Duke is higher than Duke $\rightarrow$ Market. In this last case, all backbones tend to stabilize between 20\% and 30\% of clusters removed in the last iterations. %} 

%\textcolor{red}{
What if all identities are captured by only one camera? In this extreme case, we hypothesize that the model can still adapt to the target domain. However, the performance will be limited, as different identities could be grouped in the same cluster, increasing the false positive rate. This happens because one of our assumptions is that each identity should be captured by at least two cameras. In fact, this is inherited directly from the Person Re-Identification problem. Moreover, our method utilizes this assumption to create the triplets, enabling a better adaptation to the target domain. %}

%\textcolor{red}{We faced this issue in one of the unsupervised setups in Section~\ref{subsec:impact_of_xi}. With no source pre-training, we needed to update $\xi$ for the Market dataset as this $\xi$ value was too strict, leading to clusters comprising only one camera. Consequently, all clusters were removed, and the pipeline degenerated. With $\xi = 0.03 $, the degeneration was solved as explained in Section \ref{subsec:results_unsupervised}.}  

\subsection{Impact of self-ensembling}

To check the contribution of our proposed self-ensembling method explained in Section~\ref{sec:network_fusion}, we take the best checkpoint of our model during adaptation in both scenarios, considering all backbones, and compare it with the self-ensembled model. Note that we select the best model only for reference. In practice, we do not know the best checkpoint during training since we do not have any identity-label information. Our goal here is merely to show that our self-ensembling method leads to a final model that outperforms any checkpoint individually. Even if we do not have any label information to choose the best one during training, the self-ensembling can summarize the whole training process in a final model, which is better than all checkpoints. Table~\ref{tab:ablation_fusion} shows these results.

\begin{table*}[ht]
\caption{Impact of self-ensembling. We consider a weighted average of the parameters of the backbone in different moments of the adaptation. ``Best'' refers to results obtained with the checkpoint with highest Rank-1 during adaptation. ``Fusion'' is the final model created through the proposed self-ensembling method. The best results are in \textbf{bold}.}
\label{tab:ablation_fusion}
\centering
\begin{tabular}{|P{2.5cm}|P{0.8cm}|P{1.0cm}|P{1.0cm}|P{1.0cm}|P{0.8cm}|P{1.0cm}|P{1.0cm}|P{1.0cm}|}
\hline
\multicolumn{1}{|c|}{} &
\multicolumn{4}{|c|}{Duke $\rightarrow$ Market} & \multicolumn{4}{|c|}{Market $\rightarrow$ Duke} \\
\hline
& mAP & R1 & R5 & R10 & mAP & R1 & R5 & R10 \\ \hline
ResNet (Best) & 72.1 & 89.0 & \textbf{95.5} & \textbf{97.1} & 66.2 & 81.5 & 89.5 & 92.2 \\
ResNet (Fusion) & \textbf{74.1} & \textbf{89.6} & 95.3 & \textbf{97.1} & \textbf{67.8} & \textbf{81.7} & \textbf{90.0} & \textbf{92.6} \\ \hline
OSNet (Best) & 60.7 & 85.8 & 93.5 & 95.9 & 65.1 & 81.7 & 90.3 & 92.1 \\
OSNet (Fusion) & \textbf{65.2} & \textbf{87.7} & \textbf{94.8} & \textbf{96.6} & \textbf{67.3} & \textbf{82.1} & \textbf{90.5} & \textbf{92.4} \\ \hline
DenseNet (Best)& 72.6 & 90.1 & 95.6 & 97.1 & 66.0 & 81.7 & 90.1 & 92.4\\
DenseNet (Fusion) & \textbf{76.9} & \textbf{92.0} & \textbf{96.5} & \textbf{97.7} & \textbf{69.3} & \textbf{83.4} & \textbf{91.3} & \textbf{93.0} \\\hline
\end{tabular}
\end{table*}

Our proposed self-ensembling method can improve discriminative power over the target domain by summarizing the whole training during adaptation. The method outperforms the best models in mAP by 2.0, 4.5 and 4.3 p.p., on  Duke $ \rightarrow $ Market, for ResNet50, OSNet and DenseNet121, respectively. Similarly, for Market $ \rightarrow $ Duke we achieve an improvement of 1.6, 2.2 and 3.3 p.p. in mAP for ResNet50, OSNet and DenseNet121, respectively. We can also observe gains for all backbones in both scenarios considering Rank-1. Therefore, our proposed self-ensembling strategy increases the number of correct examples retrieved from the gallery and their confidence. It shows that different checkpoints trained with different percentages of the data from the target domain have complementary information. Besides, as the self-ensembling is performed at the parameter level, without human supervision and considering each checkpoint's confidence, it reduces the memory footprint by eliminating all unnecessary checkpoints and keeping only the self-ensembled final model. 

\subsection{Impact of Ensemble-based prediction}
To increase discrimination ability, we combine distances computed by all considered architectures (Equation~\ref{eq:ensemble_backbones}) for the final inference. Results are shown in Table~\ref{tab:ablation_ensemble}.

\begin{table*}[ht]
\caption{Impact of ensemple-based prediction. Performance with and without model ensemble during inference. Best values are in \textbf{bold}.}
\label{tab:ablation_ensemble}
\centering
\begin{tabular}{|P{2.5cm}|P{0.8cm}|P{1.0cm}|P{1.0cm}|P{1.0cm}|P{0.8cm}|P{1.0cm}|P{1.0cm}|P{1.0cm}|}
\hline
\multicolumn{1}{|c|}{} &
\multicolumn{4}{|c|}{Duke $\rightarrow$ Market} & \multicolumn{4}{|c|}{Market $\rightarrow$ Duke} \\
\hline
& mAP & R1 & R5 & R10 & mAP & R1 & R5 & R10 \\ \hline
ResNet (Fusion) & 74.1 & 89.6 & 95.3 & 97.1 & 67.8 & 81.7 & 90.0 & 92.6 \\
OSNet (Fusion) & 65.2 & 87.7 & 94.8 & 96.6 & 67.3 & 82.1 & 90.5 & 92.4 \\
DenseNet (Fusion) & 76.9 & 92.0 & 96.5 & 97.7 & 69.3 & 83.4 & 91.3 & 93.0 \\
Ensembled model  & \textbf{78.4} & \textbf{92.9} & \textbf{96.9} & \textbf{97.8} & \textbf{72.6} & \textbf{85.0} & \textbf{92.1} & \textbf{93.9} \\
\hline
\end{tabular}
\end{table*}

The ensembled model outperforms the individual models by 3.3, 5.2 and 0.9 p.p. regarding Rank-1, on Duke $ \rightarrow $ Market, for ResNet50, OSNet and DenseNet, respectively. The same can be observed for Market $ \rightarrow $ Duke, in which Rank-1 is improved by 3.3, 2.9 and 1.6 p.p. for ResNet50, OSNet and DenseNet121, respectively. Results for all the other metrics also increase for both adaptation scenarios. Therefore, we can effectively combine knowledge encoded in models with different architectures. By performing it only for inference, we keep a simpler training process and still can take advantage of the ensembled knowledge from different backbones.

%(R1C2) 
%\textcolor{red}{
\subsection{Processing Footprint} %}

%\textcolor{red}{
To measure the processing footprint of our pipeline (training and inference), we consider two representative adaptation scenarios: Market $\rightarrow$ Duke and Market $\rightarrow$ MSMT17. As explained, the first setup represents a mildly difficult case and the second is the most challenging one. Table~\ref{tab:time_evaluation} shows the time measurements. %}

\begin{table*}[ht]
\caption{%\textcolor{red}{
Time Evaluation. We calculate each time in HH:MM:SS for training and in milliseconds (ms) for inference}. On training, we analyze the time taken to cluster and filter (Stages 2 and 3), one round of fine-tuning (Stage 4b), one epoch (time taken to perform $K_{2}$ iterations of orange flow), and the whole pipeline training. On inference, we calculate the time to predict the identity of a query image given the gallery feature vectors.%}
\label{tab:time_evaluation}
\centering
\begin{tabular}{|P{1.5cm}| P{1.5cm}|P{1.2cm}|P{1.0cm}|P{1.0cm}|P{1.0cm}|P{1.5cm}|P{1.2cm}|P{1.0cm}|P{1.0cm}|P{1.0cm}|}
\hline
\multicolumn{1}{|c|}{} &
\multicolumn{5}{|c|}{Market $\rightarrow$ Duke} & \multicolumn{5}{|c|}{Market $\rightarrow$ MSMT17} \\
\hline
& Clustering + filtering & Finetuning & Epoch & Whole Training & Inference & Clustering + Filtering & Finetuning & Epoch & Whole training & Inference \\ \hline
ResNet & 00:03:55 & 00:08:55 & 00:13:34 & 11:31:19 & 5ms & 00:16:45 & 00:09:36 & 00:28:08 & 23:00:55 & 13ms \\
OSNet & 00:01:53 & 00:08:56 & 00:11:14 & 09:33:04 & 4ms & 00:07:41 & 00:12:20 & 00:20:59 & 17:49:40 & 11ms\\
DenseNet & 00:04:06 & 00:08:33 & 00:13:36 & 11:33:14 & 4ms & 00:16:46 & 00:11:27 & 00:31:13 & 26:32:08 & 13ms \\
Ensemble & - & - & - & - & 6ms & - & - & - & - & 22ms
\\\hline
\end{tabular}
\end{table*}

%\textcolor{red}{
The overall time to execute the pipeline and the whole training on Market $\rightarrow$ Duke scenario is smaller than Market $\rightarrow$ MSMT17's, as expected, given that the latter is a more complex setup. %(we inform this only for reference. As explained, we assume we do not know identities on target domain). 
As the number of training images is higher, the number of proposed clusters is also higher on MSMT17. This leads to an increase in clustering, filtering, and overall training times.%}

%\textcolor{red}{
OSNet is the backbone that takes less time on both adaptation setups, because of its feature embedding size. For ResNet50 and DenseNet121, the embeddings have 2,048 dimensions while OSNet has 512. This allows a faster clustering, as Table~\ref{tab:time_evaluation} shows. Considering the same adaptation scenario, the clustering step is the most affected by the backbone and its respective embedding size. This is why ResNet50 and DenseNet121 present more similar training times and OSNet is the fastest one. %}

%\textcolor{red}{
The inference time is calculated assuming that all gallery feature vectors have been extracted and stored. It is the average time to predict the label of one query based on the ranking of the gallery images, following the protocol presented in Section~\ref{subsec:implementation_details}. The difference between both adaptation scenarios is due to the gallery size. As explained in Section~\ref{subsec:datasets}, MSMT17 has a gallery size more than $4\times$ bigger than Duke's. %}
%We do not calculate the training time on ensemble since, as we explain in Section~\ref{sec:proposed_method}, it is performed only on inference phase.}

%\textcolor{red}{
For all experiments, we used two GTX 1080 Ti GPUs. One of them is used exclusively for clustering with an implementation based on~\cite{melo2016hierarchical}, and the other for pipeline training, for each backbone. %}

\section{Conclusions and Future Work}
In this work, we tackle the problem of cross-domain Person Re-Identification (ReID) with non-overlapping cameras, especially targeting forensic scenarios with fast deployment requirements. We propose an Unsupervised Domain Adaptation (UDA) pipeline, with three novel techniques: (1) cross-camera triplet creation aiming at increasing diversity during training; (2) self-ensembling, to summarize complementary information acquired at different iterations during training; and (3) an ensemble-based prediction technique to take advantage of the complementarity between different trained backbones.

Our cross-camera triplet creation technique increases the model's invariance to different points-of-view and types of cameras in the target domain, and increases the regularization of the model, allowing the use of a single-term single-hyper-parameter triplet loss function. Moreover, we showed the importance of having this more straightforward loss function. It is less biased towards specific scenarios and helps us achieve state-of-art results in the most complex adaptation setups, surpassing prior art by a large margin in most cases. 

The self-ensembling technique helps us increase the final performance by aggregating information from different checkpoints throughout the training process, without human or label supervision. This is inspired by the reliability measurement, which shows that our models learn from more reliable data as more iterations are performed. Furthermore, this process is done in an easy-to-hard manner to increase model confidence gradually.

Finally, our last ensemble technique takes advantage of the complementarity between different backbones, enabling us to achieve state-of-the-art results without adding complexity to the training, differently from the mutual-learning strategies used in current methods ~\cite{zhai2020multiple, yin2020sskd, chen2020enhancing}. It is important to note that both ensembling strategies are done after training to generate a final model and a final prediction.

Because the training process is more straightforward than other state-of-the-art methods and does not need information on the target domain's identities, our work is easily extendable to other adaptation scenarios and deployed in actual investigations and other forensic contexts.

A key aspect of our method also shared with other recent methods in the literature~\cite{wu2019unsupervised, zhai2020ad, zhong2020learning}, is that it requires information about the camera used to acquire each sample. That is, we suppose we know, \textit{a priori}, the device that captured each image. This information does not need to be the specific type of camera but, at least, information about different camera models. Without this information, our model could face suboptimal performance, as it would not be able to take advantage of the diversity introduced by the cross-camera triplets. To address this drawback, we aim to extend this work by incorporating techniques for automatic camera attribution \cite{costa2014open, bernacki2020survey}, allowing the identification of the camera used to acquire an image or identifying whether the same camera acquired a pair of images.

Regarding the clustering process, our method requires that all selected samples are considered during this phase, which demands pairwise distance calculation between all feature vectors. Therefore, this approach may introduce higher processing times to the pipeline. In this sense, we also aim to extend our method to scale to very large datasets by introducing online deep clustering and self-supervised techniques directly in the pipeline. 

Another possible extension of our pipeline can be its application to general object re-identification, such as vehicle ReID, to mine critical objects of interest in an investigation. For example, with Person ReID, this could enable a joint analysis by matching mined identities and objects to propose relations between them during an event's analysis finally.

\section*{Acknowledgment}
We thank the financial support of the São Paulo Research Foundation (FAPESP) through the grants D\'ej\`aVu \#2017/12646-3 and \#2019/15825-1.

\ifCLASSOPTIONcaptionsoff
  \newpage
\fi



\bibliographystyle{IEEEtran}
\bibliography{IEEEabrv, refs}
%\begin{thebibliography}{1}


%\end{thebibliography}


\begin{IEEEbiography}
[{\includegraphics[width=1in,height=3cm,clip,keepaspectratio]{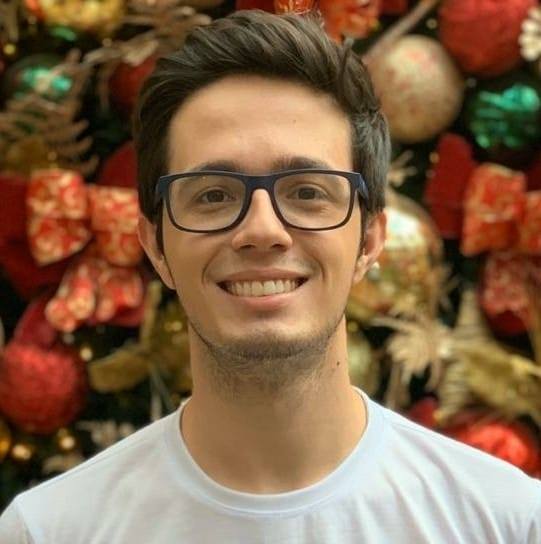}}]{Gabriel Bertocco} is currently pursuing his Ph.D. in Computer Science with a focus on digital forensics and machine learning at the Artificial Intelligence Lab. (\textbf{Recod.ai}) at the Institute of Computing, University of Campinas, Brazil, where he received a B.Sc.
in Computing Engineering in 2019. His research interests include machine learning, computer vision, and digital forensics. Contact him at gabriel.bertocco@ic.unicamp.br.
\end{IEEEbiography}

\begin{IEEEbiography}
[{\includegraphics[height=3cm,clip,keepaspectratio]{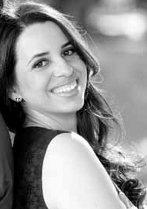}}]{Fernanda~Andal\'{o}}
is a researcher associated to the Artificial Intelligence Lab. (\textbf{Recod.ai}) at the Institute of Computing, University of Campinas, Brazil. Andal\'{o} received a Ph.D. in Computer Science from the same university in 2012, during which she was a visiting researcher at Brown University. She worked as a researcher at Samsung and as a postdoctoral researcher in collaboration with Motorola, from 2014 to 2018. Since then, she works at The LEGO Group, Denmark, devising machine learning solutions for their digital products. She is an IEEE member and was the 2016-2017 Chair of the IEEE Women in Engineering (WIE) South Brazil Section. Her research interests include machine learning and computer vision.
\end{IEEEbiography}

\begin{IEEEbiography}
[{\includegraphics[height=3cm,clip,keepaspectratio]{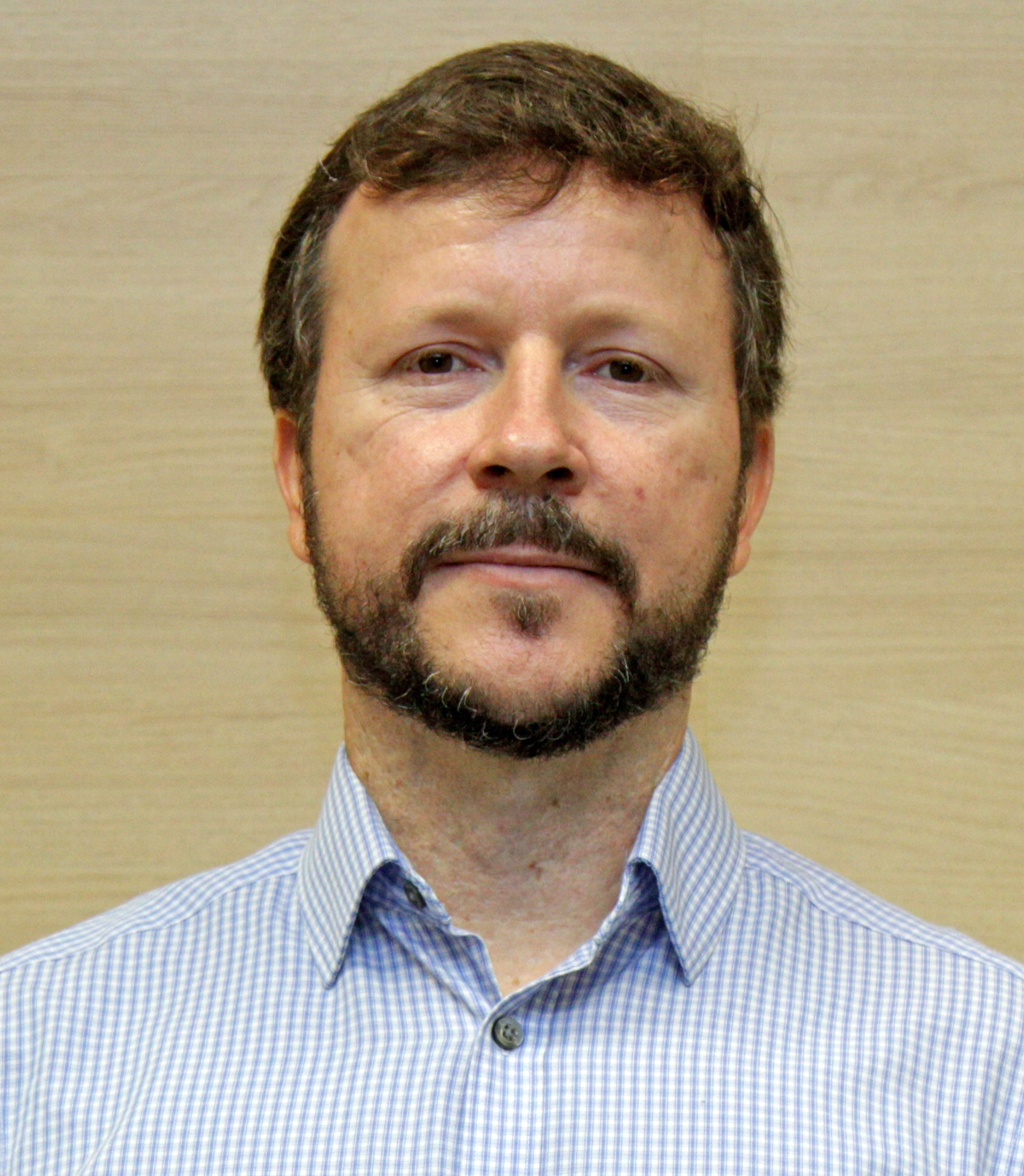}}]{Anderson Rocha} has been an associate professor
at the Institute of Computing, the University of
Campinas, Brazil, since 2009. Rocha received his Ph.D. in Computer Science from the University of Campinas. His research interests include Artificial Intelligence, Reasoning for complex data, and Digital Forensics. He is the Chair of the Artificial Intelligence Lab. (\textbf{Recod.ai}) at the Institute of Computing, University of Campinas. He was the Chair of the IEEE Information Forensics and Security Technical Committee for 2019-2020 term. Finally, Prof. is an IEEE Senior Member, a Microsoft, Google and Tan Chi Tuan Faculty Fellow and is listed among the Top-1\% of most influential scientists worldwide according to a study from Stanford Univ/Plos Biology. 
\end{IEEEbiography}


%\vfill


\end{document}

\author{Gabriel~Bertocco,
        Fernanda~Andal\'{o},~\IEEEmembership{Member,~IEEE,}\\
        and~Anderson~Rocha,~\IEEEmembership{Senior~Member,~IEEE}% <-this % stops a space
        }
\title{Supplementary Material\\\vspace{0.7cm}\Large{\textbf{Unsupervised and self-adaptative techniques for cross-domain person re-identification}}}



\markboth{IEEE Transactions on Information Forensics and Security,~Vol.~16, 2021}%
{Bertocco \MakeLowercase{\textit{et al.}}: Unsupervised and self-adaptative techniques for cross-domain person re-identification}



\maketitle

% \fi
\IEEEpeerreviewmaketitle

\section{Suplementary Material}
\label{sec:sup_mat}

\subsection{Full Comparison with Prior Art}
In the main article, we compared our results with recent state-of-the-art methods. In this supplementary material, we present the full set of results, comparing our method with methods proposed before 2018, published in top-tier journals and conferences (Tables~\ref{tab:MarketAndDuke} and~\ref{tab:msmt17}). Following the same conclusions discussed in the main article, our method outperforms all works from 2018 and previous years. 

LOMO~\cite{liao2015person}, BOW~\cite{zheng2015scalable}, and UMDL~\cite{peng2016unsupervised} are hand-crafted-based methods. They directly compute feature vectors over pixel values without using a neural network. 
UMDL also learns a shared dictionary to mine meaningful attributes from the target dataset, however, in a much simpler setup than any deep-learning method.
They then calculate the distance between query and gallery images. This makes them scalable and fast deployable. However, since hand-crafted features usually do not describe high-level features from images, the methods fail when used to match the same person from different camera views. The substantial differences caused by changes of illumination, resolution, and pose of the identities bring a high non-linearity to the feature space that is not captured by hand-crafted-based methods. We surpass UMDL by 65.3 and 66.5 percentage points (p.p) on mAP and Rank-1 when considering Market $\rightarrow$ Duke scenario and by 66.0 and 58.4 p.p. considering Duke $\rightarrow$ Market. This shows the power of deep neural networks, which effectively describe identities in a non-overlapping camera system under different point-of-views. 

All other works published in and after 2018 have been described in the main article. 
MMFA~\cite{lin2018multi} and TJ-AIDL~\cite{wang2018transferable} are methods based on low- and mid-level attribute alignment by leveraging deep convolutional neural networks. Since they do not encourage the networks to be robust to different point-of-views, their performance is lower than more recent proposed pseudo-labeling methods (PCB-PAST~\cite{zhang2019self}, SSG~\cite{fu2019self}, UDAP~\cite{song2020unsupervised}, AD-Cluster~\cite{zhai2020ad}, among others) and ensemble-based methods (ACT~\cite{yang2020asymmetric}, MMT~\cite{ge2020mutual}, MEB-Net~\cite{zhai2020multiple}, SSKD~\cite{yin2020sskd}, ABMT~\cite{chen2020enhancing}). 

The same can be observed for PTGAN~\cite{wei2018person}, SPGAN and SPGAN+LMP~\cite{deng2018image}, which are GAN-based methods that aim to transfer images from source to target domain, replicating the same camera conditions of the target domain in the labeled source images. However, transferring only camera-level features, such as color, contrast and resolution, are not enough. People on the source domain might be in different poses and contexts from the ones on the target domain, and then those methods cannot fully describe images on the target domain considering these constraints. In more recent works, researchers have proposed further processing, such as pseudo-labeling (DG-Net++~\cite{zou2020joint}), pose alignment (PDA-Net~\cite{li2019cross}), and context-alignment (CR-GAN~\cite{chen2019instance}). Our method can surpass all these GAN-based methods by a large margin. Compared to the most powerful of them, DG-Net++, we outperform it by 16.7 and 10.8 p.p on mAP and Rank-1 in the Duke $\rightarrow$ Market scenario, and in Market $\rightarrow$ Duke by 8.8 and 6.1 p.p. 

\begin{table*}[!ht]
\caption{Results on Market1501 to DukeMTMC-ReID and DukeMTMCRe-ID to Market1501 adaptation scenarios. We report mAP, Rank-1, Rank-5, and Rank-10, comparing different methods. The best result is shown in \textbf{bold}, the second in \underline{underline} and the third in \textit{italic}. Works with (*) do not pre-train the model in any source dataset before adaptation.}
\label{tab:MarketAndDuke}
\centering
\begin{tabular}{|p{2.7cm}| p{2.0cm}|p{0.8cm}|p{1.0cm}|p{1.0cm}|p{1.0cm}|p{0.8cm}|p{1.0cm}|p{1.0cm}|p{1.0cm}|}
\hline
\multicolumn{1}{|c|}{} &
\multicolumn{1}{|c|}{} &
\multicolumn{4}{|c|}{Duke $\rightarrow$ Market} & \multicolumn{4}{|c|}{Market $\rightarrow$ Duke} \\
\hline
Method & reference & mAP & R1 & R5 & R10
& mAP & R1 & R5 & R10 \\ \hline
LOMO \cite{liao2015person} & CVPR 2015 & 8.0 & 27.2 & 41.6 & 49.1 & 4.8 & 12.3 & 21.3 & 26.6 \\
BOW \cite{zheng2015scalable} & ICCV 2015 & 14.8 & 35.8 & 52.4 & 60.3 & 8.3 & 17.1 & 28.8 & 34.9 \\ 
UMDL \cite{peng2016unsupervised} & CVPR 2016 & 12.4 & 34.5 & 52.6 & 59.6 & 7.3 & 18.5 & 31.4 & 37.6 \\ \hline
PTGAN \cite{wei2018person} & CVPR 2018 & - & 38.6 & - & 66.1 & - & 27.4 & - & 50.7 \\
PUL \cite{fan2018unsupervised} & TOMM 2018 & 20.5 & 45.5 & 60.7 & 66.7 & 16.4 & 30.0 & 43.4 & 48.5\\
MMFA \cite{lin2018multi} & ArXiv & 27.4 & 56.7 & 75.0 & 81.8 & 24.7 & 45.3 & 59.8 & 66.3\\
SPGAN \cite{deng2018image} & CVPR 2018 & 22.8 & 51.5 & 70.1 & 76.8 & 22.3 & 41.1 & 56.6 & 63.0\\ 
TJ-AIDL \cite{wang2018transferable} & CVPR 2018 & 26.5 & 58.2 & 74.8 & 81.1 & 23.0 & 44.3 & 59.6 & 65.0 \\
SPGAN+LMP \cite{deng2018image} & CVPR 2018 & 26.7 & 57.7 & 75.8 & 82.4 & 26.2 & 46.4 & 62.3 & 68.0 \\
HHL \cite{zhong2018generalizing} & ECCV 2018 & 31.4 & 62.2 & 78.8 & 84.0 & 27.2 & 46.9 & 61.0 & 66.7 \\
ATNet \cite{liu2019adaptive} & CVPR 2019 & 25.6 & 55.7 & 73.2 & 79.4 & 24.9 & 45.1 & 59.5 & 64.2 \\
CamStyle \cite{zhong2018camstyle} & TIP 2019 & 27.4 & 58.8 & 78.2 & 84.3 & 25.1 & 48.4 & 62.5 & 68.9 \\
MAR \cite{yu2019unsupervised} & CVPR 2019 & 40.0 & 67.7 & 81.9 & - & 48.0 & 67.1 & 79.8 & -\\
PAUL \cite{yang2019patch} & CVPR 2019 & 40.1 & 68.5 & 82.4 & 87.4 & 53.2 & 72.0 & 82.7 & 86.0 \\
ECN \cite{zhong2019invariance} & CVPR 2019 & 43.0 & 75.1 & 87.6 & 91.6 & 40.4 & 63.3 & 75.8 & 80.4 \\
ISSDA-ReID \cite{tang2019unsupervised} & CVPR 2019 & 63.1 & 81.3 & 92.4 & 95.2 & 54.1 & 72.8 & 82.9 & 85.9 \\
PDA-Net \cite{li2019cross} & ICCV 2019 & 47.6 & 75.2 & 86.3 & 90.2 & 45.1 & 63.2 & 77.0 & 82.5 \\
CR-GAN \cite{chen2019instance} & ICCV 2019 & 54.0 & 77.7 & 89.7 & 92.7 & 48.6 & 68.9 & 80.2 & 84.7 \\
PCB-PAST \cite{zhang2019self} & ICCV 2019 & 54.6 & 78.4 & - & - & 54.3 & 72.4 & - & - \\
UCDA \cite{qi2019novel} & ICCV 2019 & 30.9 & 60.4 & - & - & 31.0 & 47.7 & - & - \\
%\textcolor{red}{SSG \cite{fu2019self}} & \textcolor{red}{ICCV 2019} & \textcolor{red}{58.3} & \textcolor{red}{80.0}& \textcolor{red}{90.0} & \textcolor{red}{92.4} & \textcolor{red}{53.4} & \textcolor{red}{73.0} & \textcolor{red}{80.6} & \textcolor{red}{83.2} \\
SSG \cite{fu2019self} & ICCV 2019 & 58.3 & 80.0 & 90.0 & 92.4 & 53.4 & 73.0 & 80.6 & 83.2 \\
CASCL \cite{wu2019unsupervised} & ICCV 2019 & 35.5 & 65.4 & 80.6 & 86.2 & 37.8 & 59.3 & 73.2 & 77.8 \\
SSL \cite{lin2020unsupervised}* & CVPR 2020 & 37.8 & 71.7 & 83.8 & 87.4 & 28.6 & 52.5 & 63.5 & 68.9 \\
%\textcolor{red}{CCSE \cite{lin2020unsupervisedccse}*} & \textcolor{red}{TIP 2020} & \textcolor{red}{38.0} & \textcolor{red}{73.7} & \textcolor{red}{84.0} & \textcolor{red}{87.9} & \textcolor{red}{30.6}& \textcolor{red}{56.1} & \textcolor{red}{66.7} & \textcolor{red}{71.5} \\ 
CCSE \cite{lin2020unsupervisedccse}* & TIP 2020 & 38.0 & 73.7 & 84.0 & 87.9 & 30.6 & 56.1 & 66.7 & 71.5 \\ 
UDAP \cite{song2020unsupervised} & PR 2020 & 53.7 & 75.8 & 89.5 & 93.2 & 49.0 & 68.4 & 80.1 & 83.5 \\
MMCL \cite{wang2020unsupervised}& CVPR 2020 & 60.4 & 84.4 & 92.8 & 95.0 & 51.4 & 72.4 & 82.9 & 85.0 \\
ACT \cite{yang2020asymmetric} & AAAI 2020 & 60.6 & 80.5 & - & - & 54.5 & 72.4 & - & - \\
ECN-GPP \cite{zhong2020learning} & TPAMI 2020 & 63.8 & 84.1 & 92.8 & 95.4 & 54.4 & 74.0 & 83.7 & 87.4 \\
HCT \cite{zeng2020hierarchical}* & CVPR 2020 & 56.4 & 80.0 & 91.6 & 95.2 & 50.7 & 69.6 & 83.4 & 87.4 \\ 
SNR \cite{jin2020style} & CVPR 2020 & 61.7 & 82.8 & - & - & 58.1 & 76.3 & - & - \\
AD-Cluster \cite{zhai2020ad} & CVPR 2020 & 68.3 & 86.7 & 94.4 & 96.5 & 54.1 & 72.6 & 82.5 & 85.5 \\
MMT \cite{ge2020mutual} & ICLR 2020 & 71.2 & 87.7 & 94.9 & 96.9 & 65.1 & 78.0 & 88.8 & \textit{92.5} \\
CycAs \cite{wang2020cycas}* & ECCV 2020 & 64.8 & 84.8 & - & - & 60.1 & 77.9 & - & - \\
DG-Net++ \cite{zou2020joint} & ECCV 2020 & 61.7 & 82.1 & 90.2 & 92.7 & 63.8 & 78.9 & 87.8 & 90.4 \\
MEB-Net \cite{zhai2020multiple} & ECCV 2020 & 76.0 & 89.9 & 96.0 & 97.5 & 66.1 & 79.6 & 88.3 & 92.2 \\
Dual-Refinement \cite{dai2020dual} & arXiv 2020 & 78.0 & 90.9 & \textit{96.4} & \textit{97.7} & 67.7 & 82.1 & 90.1 & \textit{92.5} \\
SSKD \cite{yin2020sskd} & arXiv 2020 & \textit{78.7} & 91.7 & \textbf{97.2} & \textbf{98.2} & 67.2 & 80.2 & \textit{90.6} & \underline{93.3} \\ 
ABMT \cite{chen2020enhancing} & WACV 2020 & \underline{80.4} & \underline{93.0} & - & - & \textit{70.8} & \textit{83.3} & - & - \\
\hline
%\textcolor{red}{\textbf{Ours (w/o Re-Ranking)*}} & \textcolor{red}{This Work} & \textcolor{red}{67.7} & \textcolor{red}{89.5}  & \textcolor{red}{94.8} & \textcolor{red}{96.5} & \textcolor{red}{68.8} & \textcolor{red}{82.4} & \textcolor{red}{\textit{90.6}} & \textcolor{red}{\textit{92.5}} \\ \hline
\textbf{Ours (w/o Re-Ranking)*} & This Work & 67.7 89.5  & 94.8 & 96.5 & 68.8 & 82.4 & \textit{90.6} & \textit{92.5} \\ \hline
\textbf{Ours (w/o Re-Ranking)} & This Work & 78.4 & \textit{92.9} & \underline{96.9} & \underline{97.8} & \underline{72.6} & \underline{85.0} & \underline{92.1} & \textbf{93.9} \\ \hline
\textbf{Ours (w/ Re-Ranking)} & This Work & \textbf{88.0} & \textbf{93.8} & \textit{96.4} & 97.4 & \textbf{82.7} & \textbf{87.2} & \textbf{92.5} & \textbf{93.9} \\ \hline
\end{tabular}
\end{table*}

MSMT17 is the most recent and challenging Person ReID benchmark, and this is why there are fewer works considering it for evaluation. PTGAN was the first to considered it, both being proposed in~\cite{wei2018person}, in 2018.
We outperform them by a substantial margin of 31.2 and 52.1 p.p. in mAP and Rank-1 in Duke $\rightarrow$ MSMT17 and by 30.3 and 52.1 p.p in the challenging Market $\rightarrow$ MSMT17 scenario. 

SpCL~\cite{ge2020self} is similar to ours in the sense that it increases the cluster reliability during the clustering stage as the training progresses. However, it does not apply any strategy considering diversity as we do by creating diverse triplets considering all cameras comprised in a cluster. Besides, they leverage both source and target domain images on adaptation stages and enable their model to use the source labeled identity to bring some regularization to the adaptation process. Differently, our method does not use the source domain images after fine-tuning and leverages the adaptation process relying only on target images. As shown in the main article, we outperform them by 1.2 and 4.2 p.p. in the most challenging Market $\rightarrow$ MSMT17 in mAP and Rank-1, respectively.      

\begin{table*}[ht]
\caption{Results on Market1501 to MSMT17 and DukeMTMCRe-ID to MSMT17 adaptation scenarios. We report mAP, Rank-1, Rank-5, and Rank-10, comparing different methods. The best result is shown in \textbf{bold}, the second in \underline{underline} and the third in \textit{italic}. Works with (*) do not pre-train the model in any source dataset before adaptation.}
\label{tab:msmt17}
\centering
\begin{tabular}{|p{2.7cm}| p{2.0cm}|p{0.8cm}|p{1.0cm}|p{1.0cm}|p{1.0cm}|p{0.8cm}|p{1.0cm}|p{1.0cm}|p{1.0cm}|}
\hline
\multicolumn{1}{|c|}{} &
\multicolumn{1}{|c|}{} &
\multicolumn{4}{|c|}{Duke $\rightarrow$ MSMT17} & \multicolumn{4}{|c|}{Market $\rightarrow$ MSMT17} \\
\hline
Method & reference & mAP & R1 & R5 & R10
& mAP & R1 & R5 & R10 \\ \hline
PTGAN \cite{wei2018person} & CVPR 2018 & 3.3 & 11.8 & - & 27.4 & 2.9 & 10.2 & - & 24.4 \\
ECN \cite{zhong2019invariance} & CVPR 2019 & 10.2 & 30.2 & 41.5 & 46.8 & 8.5 & 25.3 & 36.3 & 42.1 \\
%\textcolor{red}{CCSE \cite{lin2020unsupervisedccse}*} & \textcolor{red}{TIP 2020} & \textcolor{red}{9.9} & \textcolor{red}{31.4} & \textcolor{red}{41.4} & \textcolor{red}{45.7} & \textcolor{red}{9.9} & \textcolor{red}{31.4} & \textcolor{red}{41.4} & \textcolor{red}{45.7} \\
%\textcolor{red}{SSG \cite{fu2019self}} & \textcolor{red}{ICCV 2019} & \textcolor{red}{13.3} & \textcolor{red}{32.2} & \textcolor{red}{-} & \textcolor{red}{51.2} & \textcolor{red}{13.2} & \textcolor{red}{31.6} & \textcolor{red}{-} & \textcolor{red}{49.6} \\
CCSE \cite{lin2020unsupervisedccse}* & TIP 2020 & 9.9 & 31.4 & 41.4 & 45.7 & 9.9 & 31.4 & 41.4 & 45.7 \\
SSG \cite{fu2019self} & ICCV 2019 & 13.3 & 32.2 & - & 51.2 & 13.2 & 31.6 & - & 49.6 \\
ECN-GPP \cite{zhong2020learning} & TPAMI 2020 & 16.0 & 42.5 & 55.9 & 61.5 & 15.2 & 40.4 & 53.1 & 58.7 \\
MMCL \cite{wang2020unsupervised} & CVPR 2020 & 16.2 & 43.6 & 54.3 & 58.9 & 15.1 & 40.8 & 51.8 & 56.7 \\
MMT \cite{ge2020mutual} & ICLR 2020 & 23.3 & 50.1 & 63.9 & 69.8 & 22.9 & 49.2 & 63.1 & 68.8 \\
CycAs \cite{wang2020cycas}* & ECCV 2020 & 26.7 & 50.1 & - & - & 26.7 & 50.1 & - & - \\
DG-Net++  \cite{zou2020joint} & ECCV 2020 & 22.1 & 48.8 & 60.9 & 65.9 & 22.1 & 48.4 & 60.9 & 66.1 \\
Dual-Refinement \cite{dai2020dual} & arXiv 2020 & 26.9 & 55.0 & 68.4 & 73.2 & 25.1 & 53.3 & 66.1 & 71.5 \\
SSKD \cite{yin2020sskd} & arXiv 2020 & 26.0 & 53.8 & \textit{66.6} & \textit{72.0} & 23.8 & 49.6 & 63.1 & 68.8 \\ 
ABMT \cite{chen2020enhancing} & WACV 2020 & \textit{33.0} & \textit{61.8} & - & - & 27.8 & 55.5 & - & - \\ 
SpCL \cite{ge2020self} & NeurIPS 2020 & - & - & - & - & \textit{31.0} & \textit{58.1} & \textit{69.6} & \textit{74.1} \\ \hline
\textbf{Ours (w/o Re-Ranking)} & This Work & \underline{34.5} & \underline{63.9} & \underline{75.3} & \underline{79.6} & \underline{33.2} & \underline{62.3} & \underline{74.1} & \underline{78.5} \\ \hline
\textbf{Ours (w/ Re-Ranking)} & This Work & \textbf{46.6} & \textbf{69.6} & \textbf{77.1} & \textbf{80.4} & \textbf{45.2} & \textbf{68.1} & \textbf{76.0} & \textbf{79.2} \\ \hline
\end{tabular}
\end{table*}

%\textcolor{red}{
\subsection{%(R1C3) - 
Full Qualitative Analysis} %}

%\textcolor{red}{
In this section we extend the qualitative analysis presented in the main article. We show the successful and failure cases for a query from each camera for Market1501 and DukeMTMC-ReID datasets. %} 

\begin{figure*}[ht]
\centering
\subfloat[Camera 1]{\includegraphics[width=3.5in]{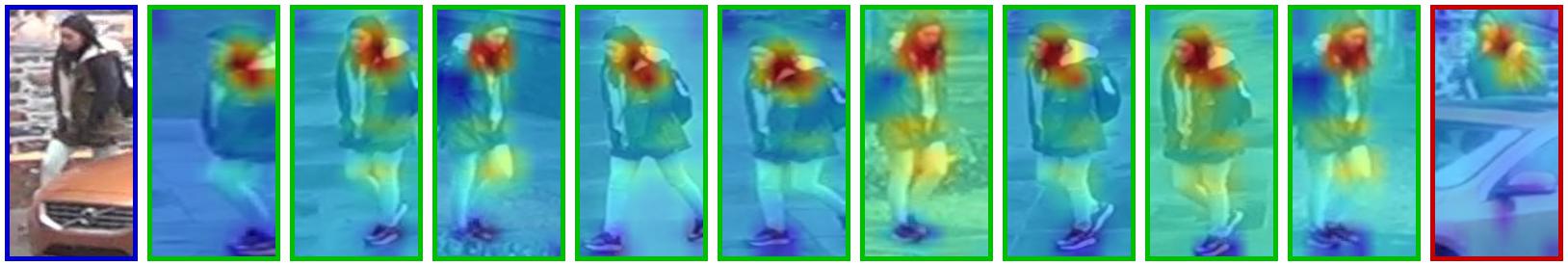}
\label{fig:M2D_successful_cam01}}
\hfil
\subfloat[Camera 2]{\includegraphics[width=3.5in]{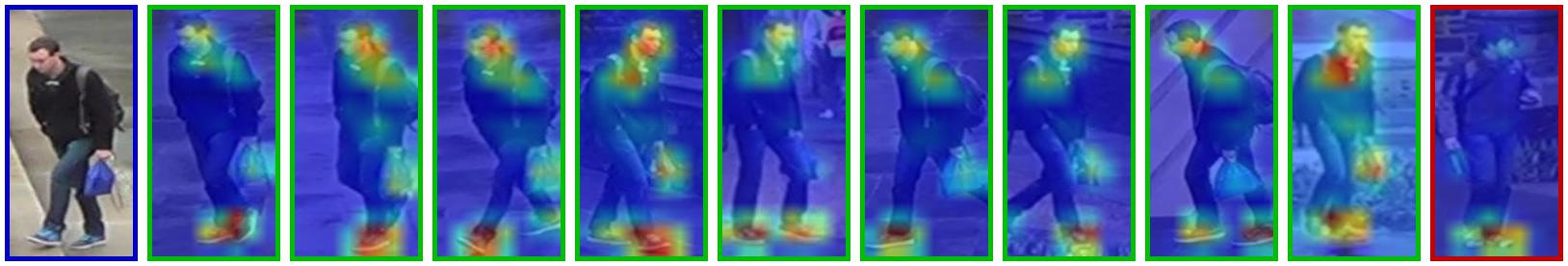} 
\label{fig:M2D_successful_cam02}}
\hfil
\subfloat[Camera 3]{\includegraphics[width=3.5in]{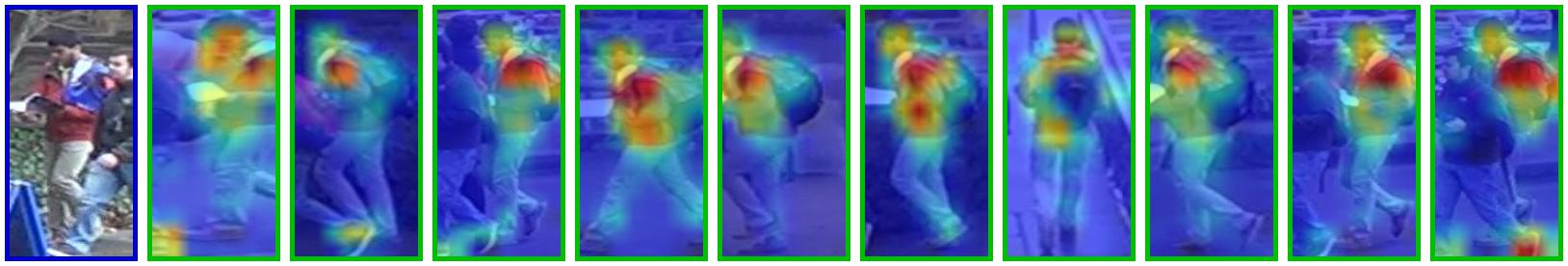} 
\label{fig:M2D_successful_cam03}}
\hfil
\subfloat[Camera 4]{\includegraphics[width=3.5in]{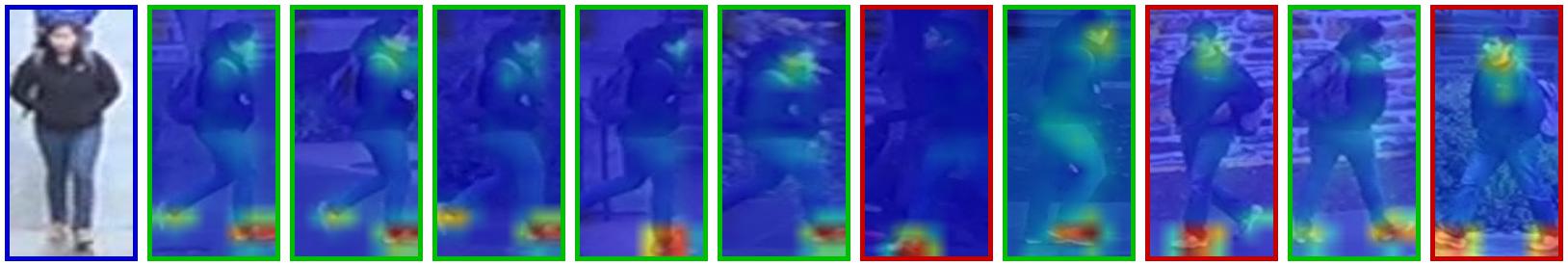} 
\label{fig:M2D_successful_cam04}}
\hfil
\subfloat[Camera 5]{\includegraphics[width=3.5in]{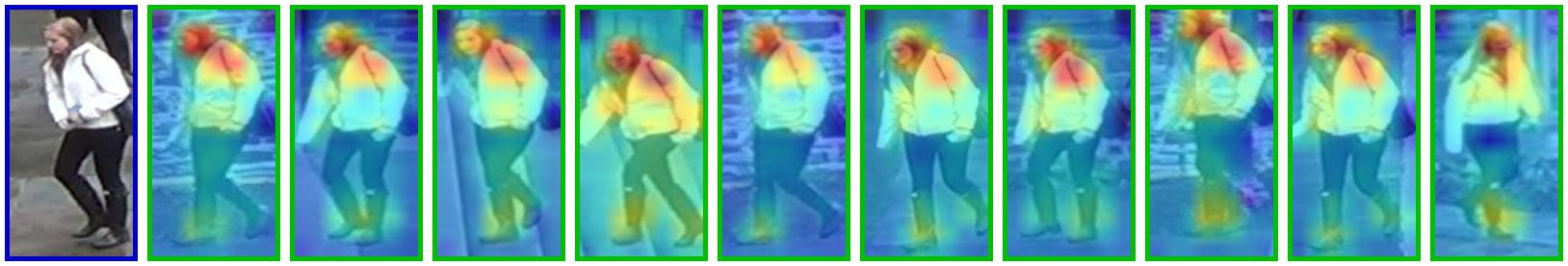} 
\label{fig:M2D_successful_cam05}}
\hfil
\subfloat[Camera 6]{\includegraphics[width=3.5in]{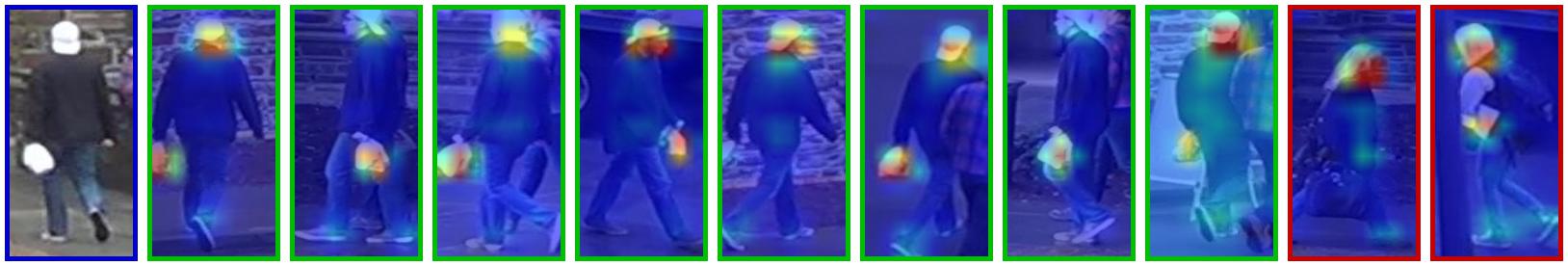} 
\label{fig:M2D_successful_cam06}}
\hfil
\subfloat[Camera 7]{\includegraphics[width=3.5in]{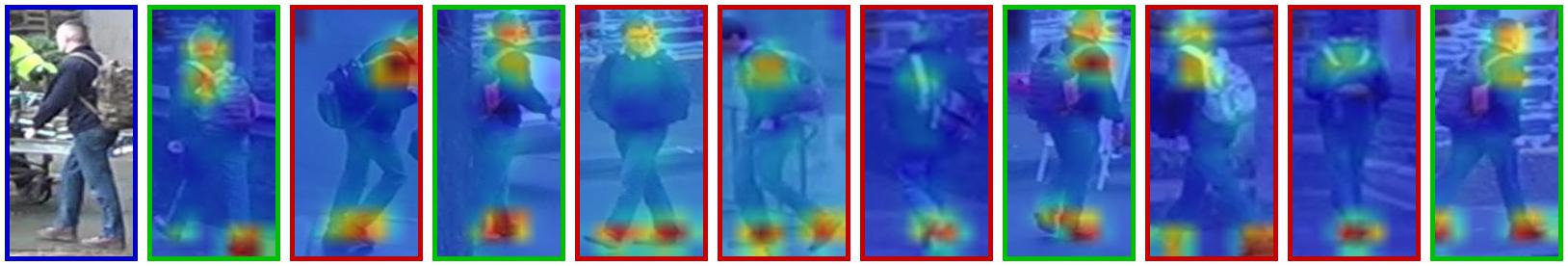} 
\label{fig:M2D_successful_cam07}}
\hfil
\subfloat[Camera 8]{\includegraphics[width=3.5in]{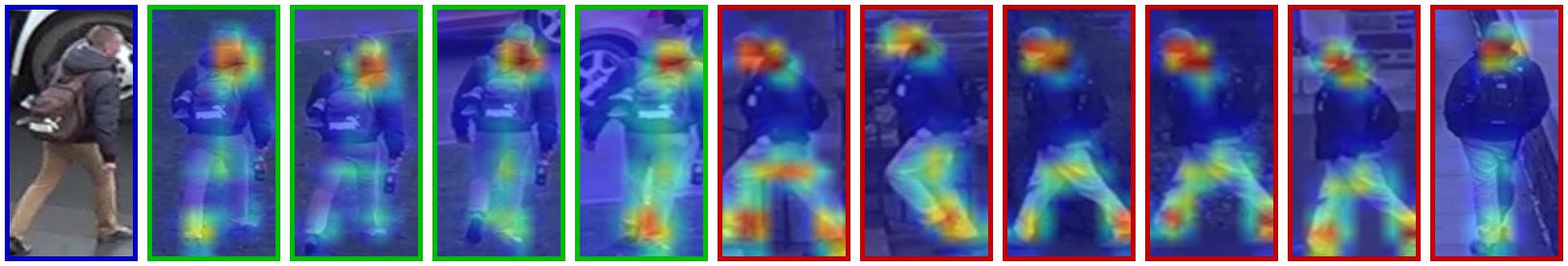} 
\label{fig:M2D_successful_cam08}}
\caption{Successful cases considering one query from each camera on Duke. This results are obtained with the ResNet50 backbone after the Market $\rightarrow$ Duke adaptation.}
\label{fig:M2D_successful_cases}
\end{figure*}

\begin{figure*}[ht]
\centering
\subfloat[Camera 1]{\includegraphics[width=3.5in]{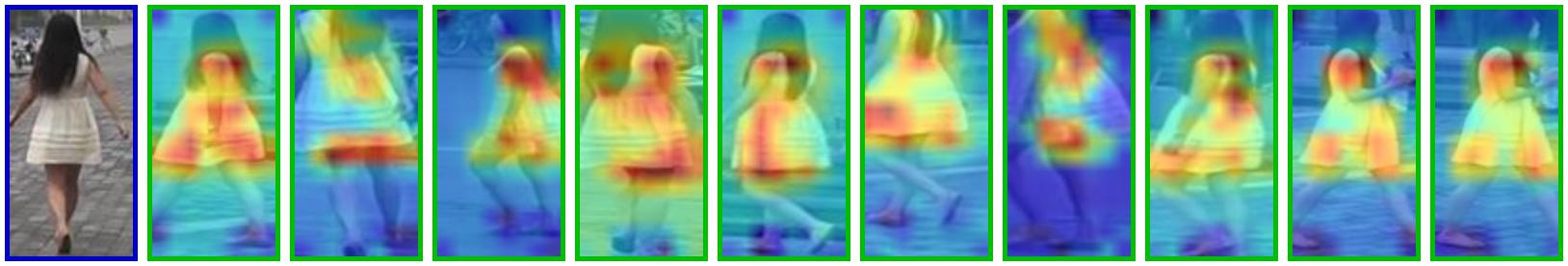}
\label{fig:D2M_successful_cam01}}
\hfil
\subfloat[Camera 2]{\includegraphics[width=3.5in]{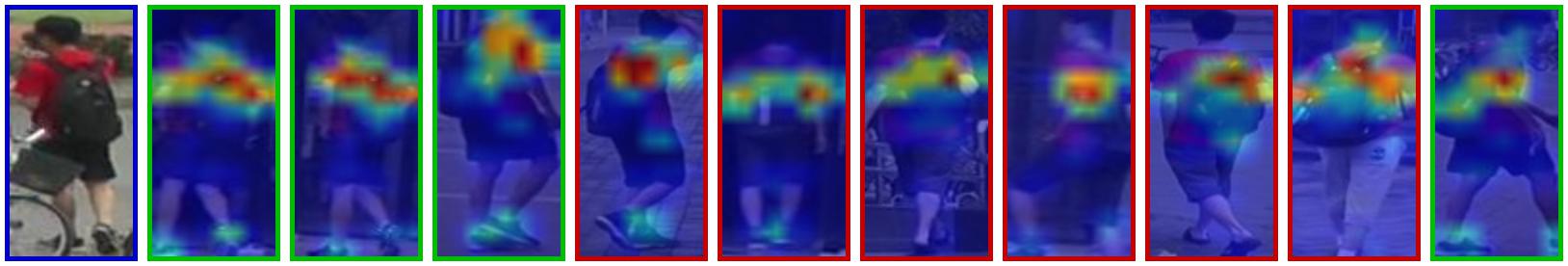} 
\label{fig:D2M_successful_cam02}}
\hfil
\subfloat[Camera 3]{\includegraphics[width=3.5in]{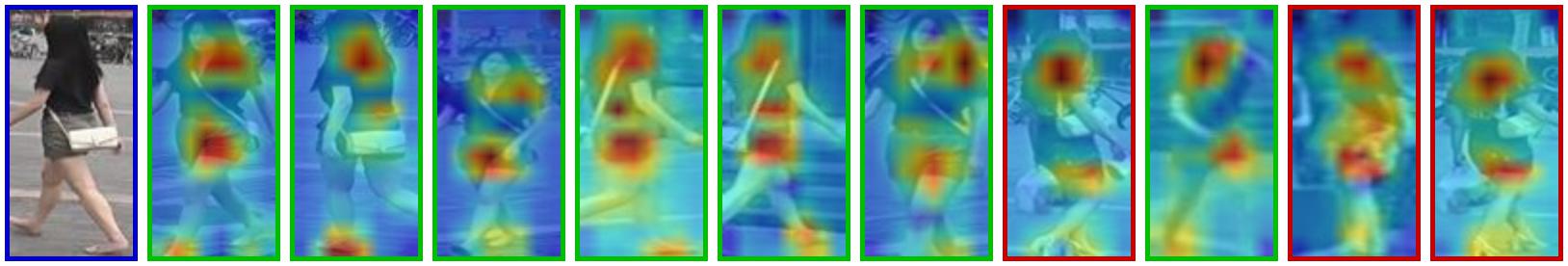} 
\label{fig:D2M_successful_cam03}}
\hfil
\subfloat[Camera 4]{\includegraphics[width=3.5in]{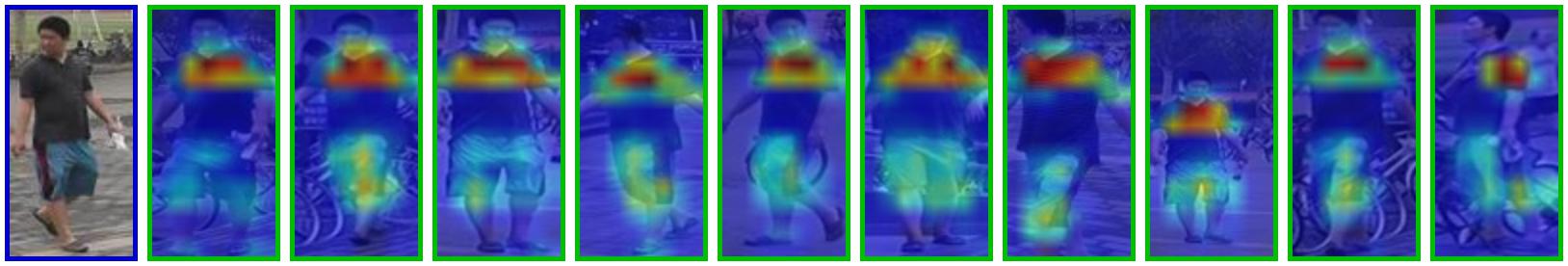} 
\label{fig:D2M_successful_cam04}}
\hfil
\subfloat[Camera 5]{\includegraphics[width=3.5in]{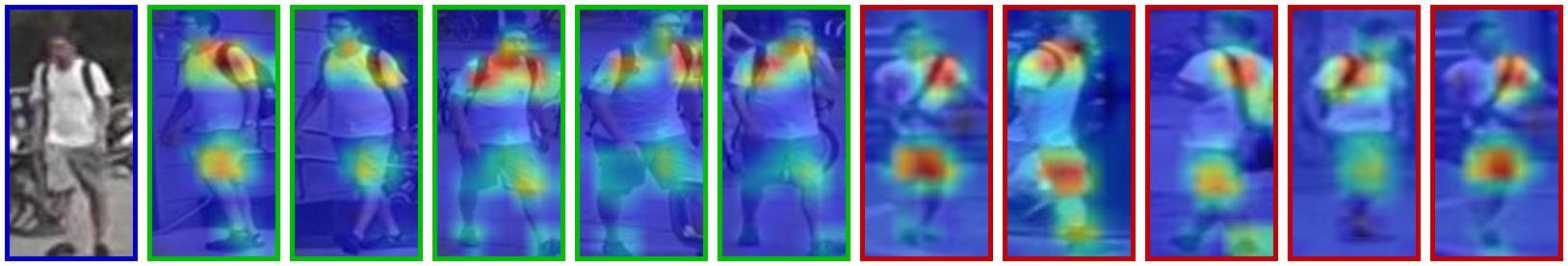} 
\label{fig:D2M_successful_cam05}}
\hfil
\subfloat[Camera 6]{\includegraphics[width=3.5in]{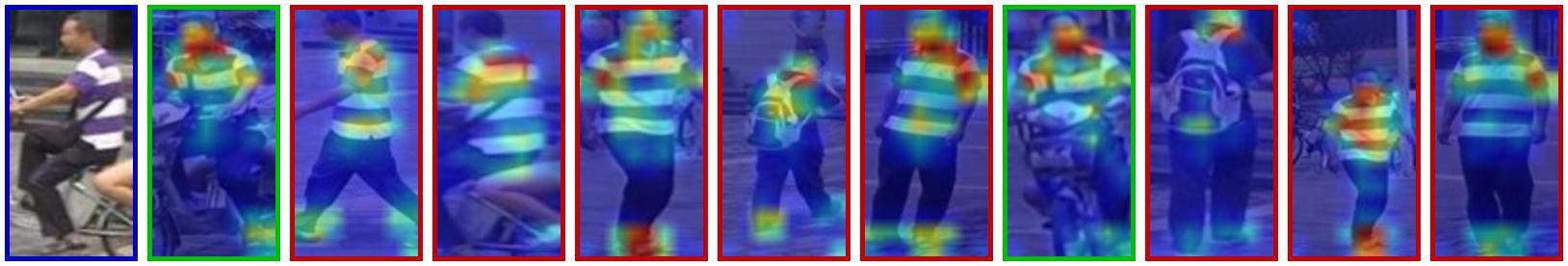} 
\label{fig:D2M_successful_cam06}}
\caption{Successful cases considering one query from each camera on Market. This results are obtained with the ResNet50 backbone after the Duke $\rightarrow$ Market adaptation.}
\label{fig:D2M_successful_cases}
\end{figure*}

%\textcolor{red}{
In Figures~\ref{fig:M2D_successful_cases} and~\ref{fig:D2M_successful_cases}, we observe some successful cases with the activation maps for the top-10 closest gallery images to the query. We adapted the implementation from~\cite{stylianou2019visualizing} to visualize the activation maps. In both scenarios, we see that our model is able to find fine-grained details on the images, enabling it to correct match the query to the gallery images. For example, in Figure~\ref{fig:M2D_successful_cam04}, the model tends to focus on shoes and parts of the face, while in Figure~\ref{fig:M2D_successful_cam01} the focus is on regions depicting hair and pants. We conclude that our method is able to distinguish the semantic parts of the body and soft-biometric attributes which are vital to Person Re-Identification. It is also important to note that the query and the correct matches are always from different cameras, which also confirms that our model is able to overcome different camera conditions. Analyzing the errors in Figures~\ref{fig:M2D_successful_cam08} and~\ref{fig:D2M_successful_cam06}, we see they are mainly caused by similar clothes, but the method is still able to recover at least the closest gallery image (Rank-1 image). %}

\begin{figure*}[ht]
\centering
\subfloat[Camera 1]{\includegraphics[width=3.5in]{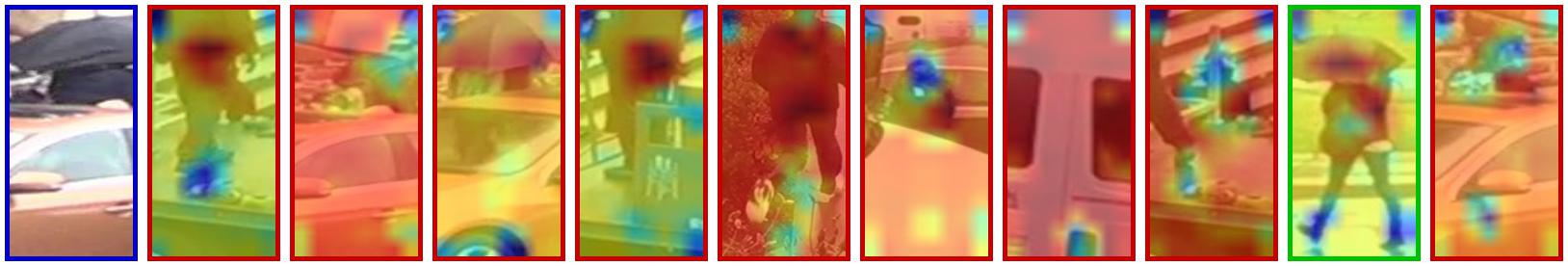}
\label{fig:M2D_failure_cam01}}
\hfil
\subfloat[Camera 2]{\includegraphics[width=3.5in]{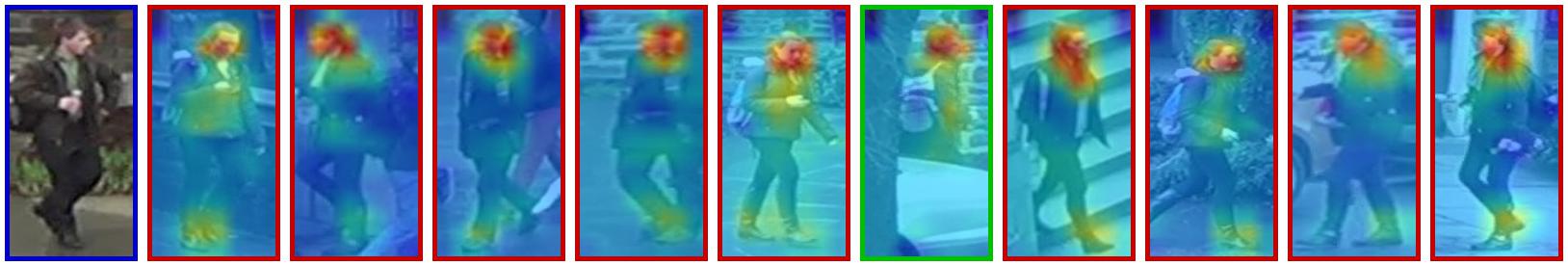} 
\label{fig:M2D_failure_cam02}}
\hfil
\subfloat[Camera 3]{\includegraphics[width=3.5in]{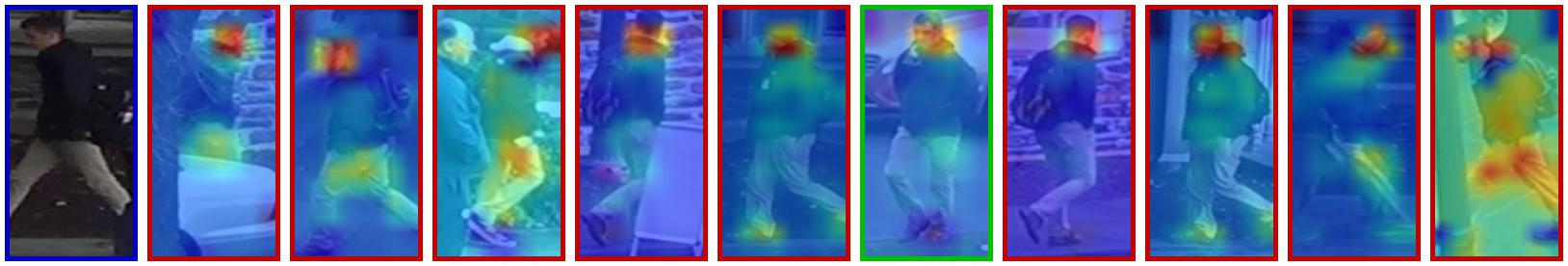} 
\label{fig:M2D_failure_cam03}}
\hfil
\subfloat[Camera 4]{\includegraphics[width=3.5in]{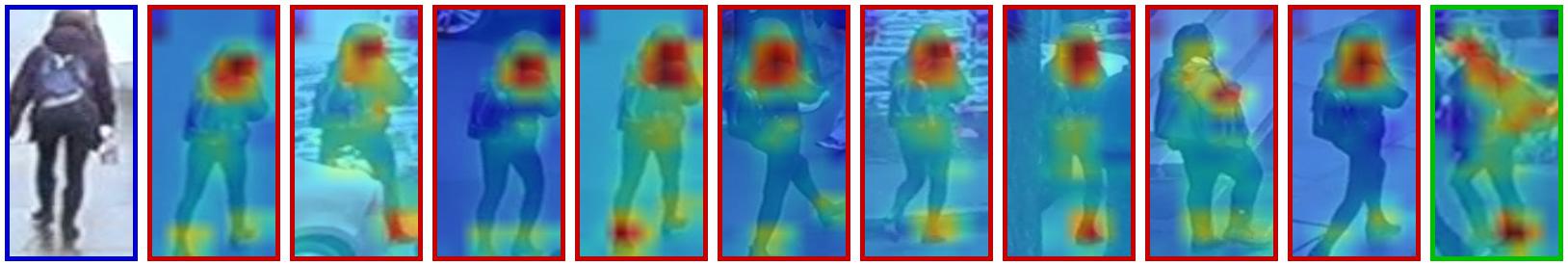} 
\label{fig:M2D_failure_cam04}}
\hfil
\subfloat[Camera 5]{\includegraphics[width=3.5in]{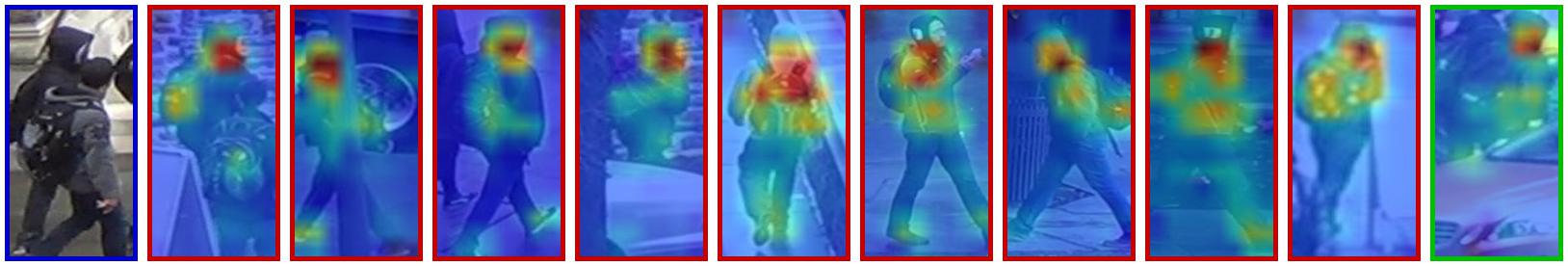} 
\label{fig:M2D_failure_cam05}}
\hfil
\subfloat[Camera 6]{\includegraphics[width=3.5in]{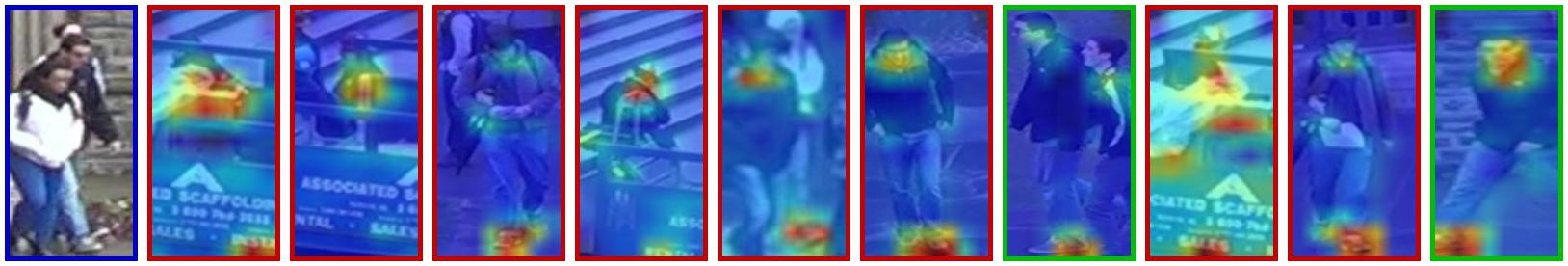} 
\label{fig:M2D_failure_cam06}}
\hfil
\subfloat[Camera 7]{\includegraphics[width=3.5in]{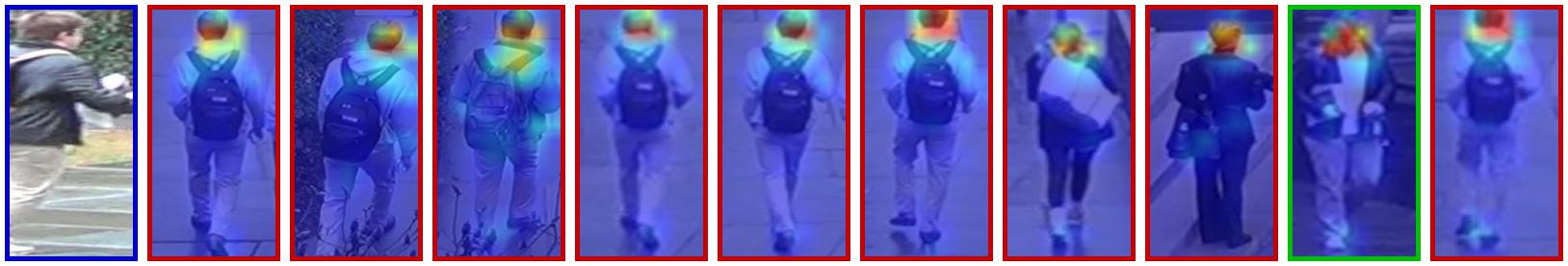} 
\label{fig:M2D_failure_cam07}}
\hfil
\subfloat[Camera 8]{\includegraphics[width=3.5in]{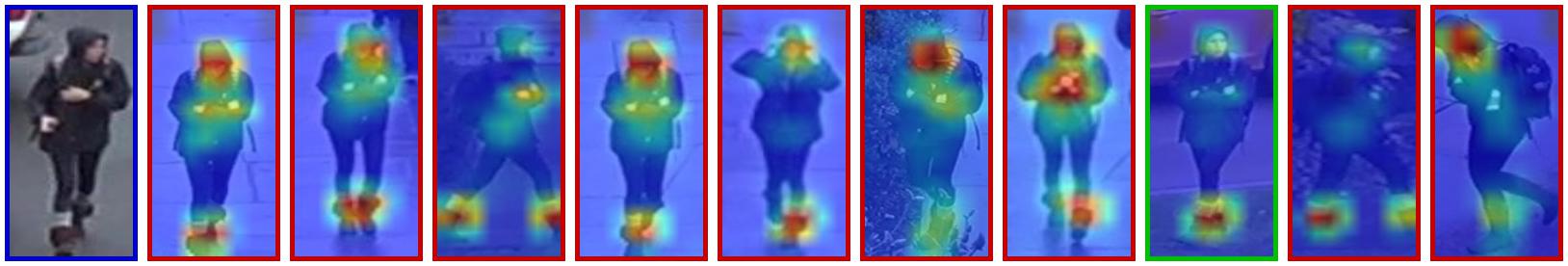} 
\label{fig:M2D_failure_cam08}}
\caption{Failure cases considering one query from each camera on Duke. This results are obtained with the ResNet50 backbone after the Market $\rightarrow$ Duke adaptation.}
\label{fig:M2D_failure_cases}
\end{figure*}

\begin{figure*}[ht]
\centering
\subfloat[Camera 1]{\includegraphics[width=3.5in]{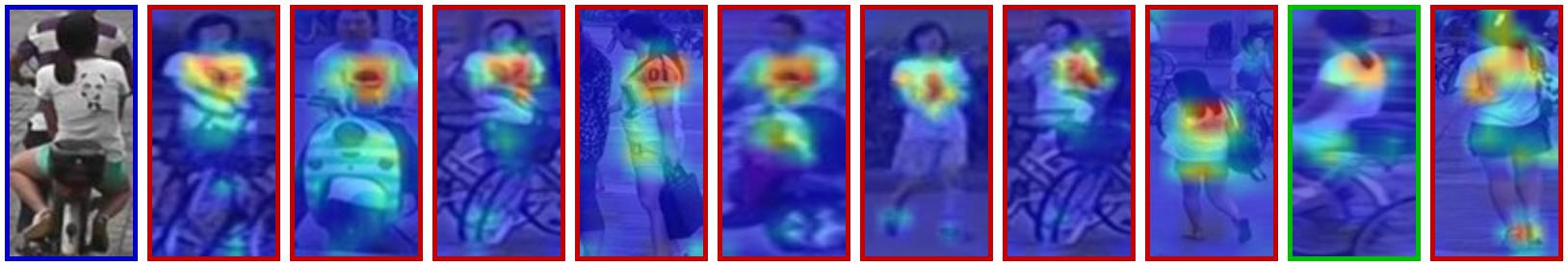}
\label{fig:D2M_failure_cam01}}
\hfil
\subfloat[Camera 2]{\includegraphics[width=3.5in]{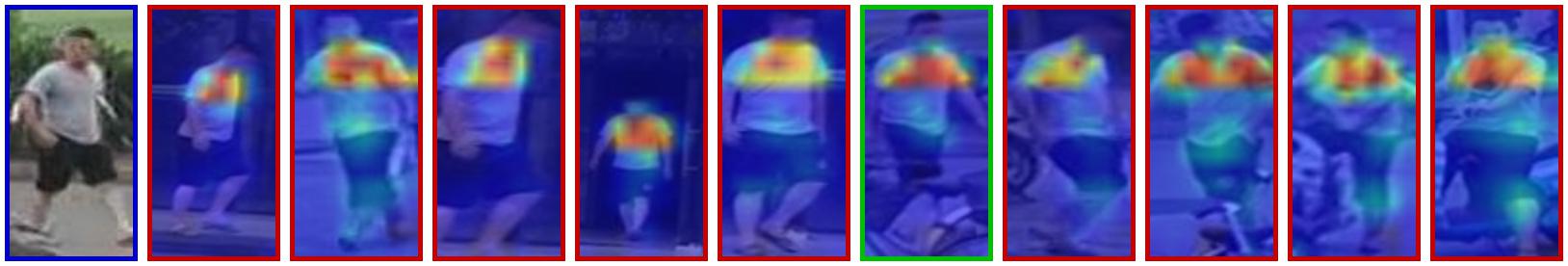} 
\label{fig:D2M_failure_cam02}}
\hfil
\subfloat[Camera 3]{\includegraphics[width=3.5in]{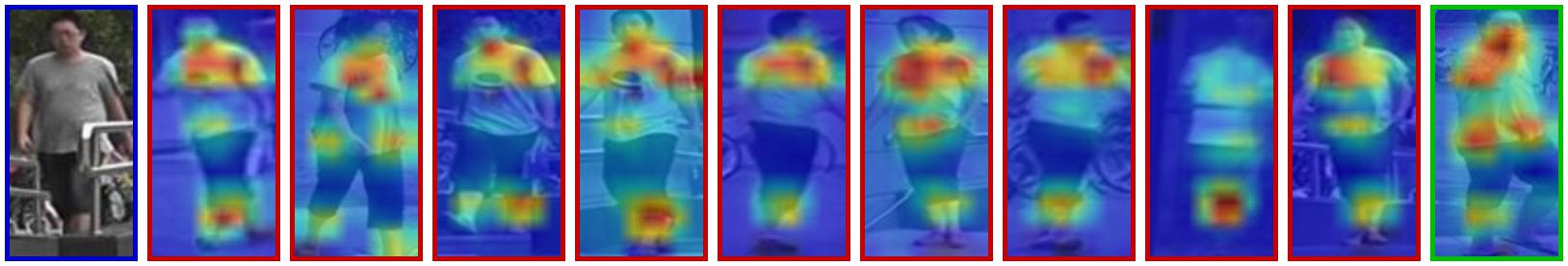} 
\label{fig:D2M_failure_cam03}}
\hfil
\subfloat[Camera 4]{\includegraphics[width=3.5in]{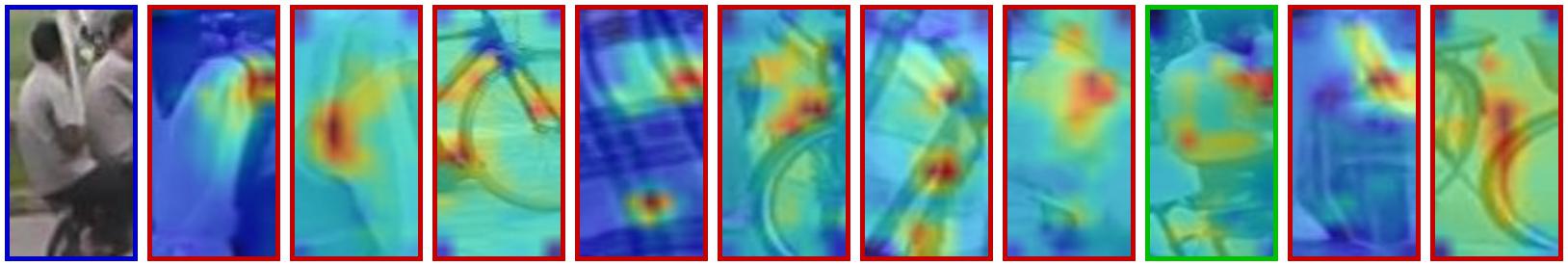} 
\label{fig:D2M_failure_cam04}}
\hfil
\subfloat[Camera 5]{\includegraphics[width=3.5in]{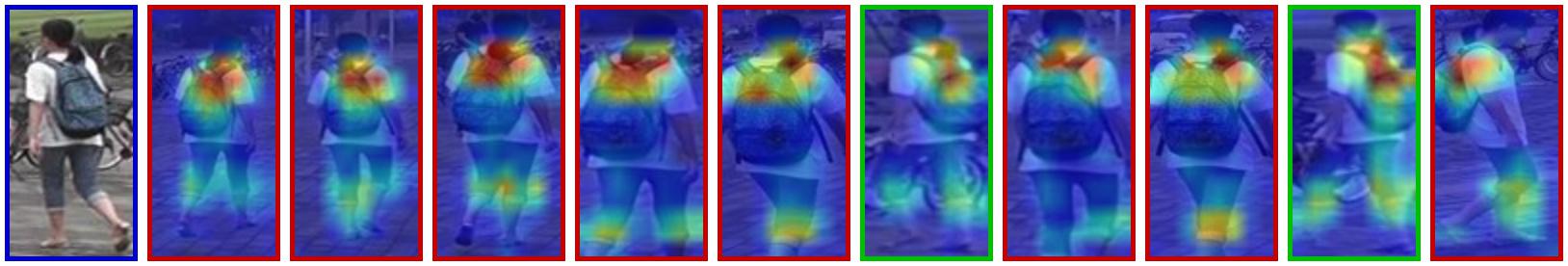} 
\label{fig:D2M_failure_cam05}}
\hfil
\subfloat[Camera 6]{\includegraphics[width=3.5in]{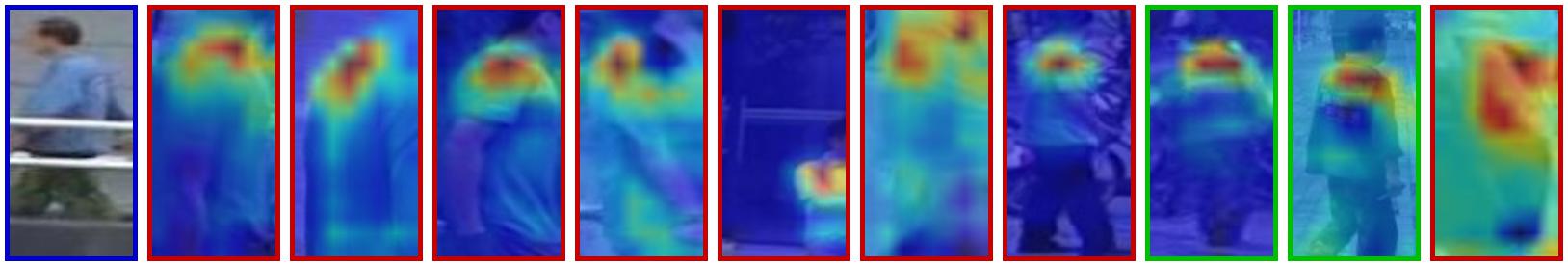} 
\label{fig:D2M_failure_cam06}}
\caption{Failure cases considering one query from each camera on Market. This results are obtained with the ResNet50 backbone after the Duke $\rightarrow$ Market adaptation.}
\label{fig:D2M_failure_cases}
\end{figure*}

%\textcolor{red}{
Figures~\ref{fig:M2D_failure_cases} and~\ref{fig:D2M_failure_cases} depict the failure cases for one of the backbones, showing the limitations of the method. The failures are mainly related to similar clothes or soft-biometric attributes. Despite the high probability of people wearing similar clothing items, it is marginal the chance of them being dressed exactly the same way, however this is the situation shown in Figure~\ref{fig:D2M_failure_cam02}.%}

%\textcolor{red}{
Another source of errors is occlusion. Figure~\ref{fig:M2D_failure_cam01} is an example where the person has been fully occluded by a car. 
As there is no person, the method does not have any specific region to focus on and then the gallery images are almost fully activated. In Figure~\ref{fig:D2M_failure_cam04}, the target identity is on a motorcycle together with another person, which led them to be in the same bounding box. In this case, the method erroneously retrieves images with no identity on it (distractor images on gallery) or images with parts of a bike. %}

%\textcolor{red}{
Figure~\ref{fig:D2M_failure_cam01} shows an interesting failure case, where the model focus uniquely on the drawing on the person's shirt in the query image. The method returns gallery images of other identities with similar shirt drawings. Despite the failure, it is interesting to note that our method was able to focus on fine-grained details to find matches, and not activate the whole image or large parts of it.%} 

% or


\ifCLASSOPTIONcaptionsoff
  \newpage
\fi



\bibliographystyle{IEEEtran}
\bibliography{IEEEabrv, refs}
%\begin{thebibliography}{1}


%\end{thebibliography}



%\vfill

